# From bits to atoms: A survey of cross-layer safety in Embodied AI

Tuo Feng,[1] Yue Zhang,[2,*] Shiji Zhou,[3,*] Zhaoxin Fan,[3,*] Wenguan Wang,[1,*] Xiaowei Chi,[4] Tianyu Shen,[5] Xiangtai Li,[6] Xuesong Bai,[3,7] Zheng Zhu,[8] Sirui Han,[4] Boyu Zhou,[9,10] Zhaosong Wang,[11] Shizhe Chen,[12] Zilong Zheng,[13] Chenjia Bai,[14] Jiangmiao Pang,[15] Yaodong Yang,[16] Hao Zhao,[17] Di Hu,[18] Li Yi,[17] Yan Huang,[19,20] Qi Wu,[21] Hanwang Zhang,[6] Fei Gao,[1,10] Shanghang Zhang,[16] Hongsheng Li,[22] Jianye Hao,[23] Yuan Zhang,[24] Yingjie Xia,[25] Jian Shen,[26] Cewu Lu,[27] Hui Zhang,[28] Kai Chen,[29] Wenwu Yu,[30] Badong Chen,[31] Jungong Han,[17] and Yike Guo[4]
*Correspondence: zyueinfosec@sdu.edu.cn (Y.Z.); zhoushiji25@buaa.edu.cn (S.Z.); zhaoxinf@buaa.edu.cn (Z.F.); wenguanwang@zju.edu.cn (W.W.)

## GRAPHICAL ABSTRACT

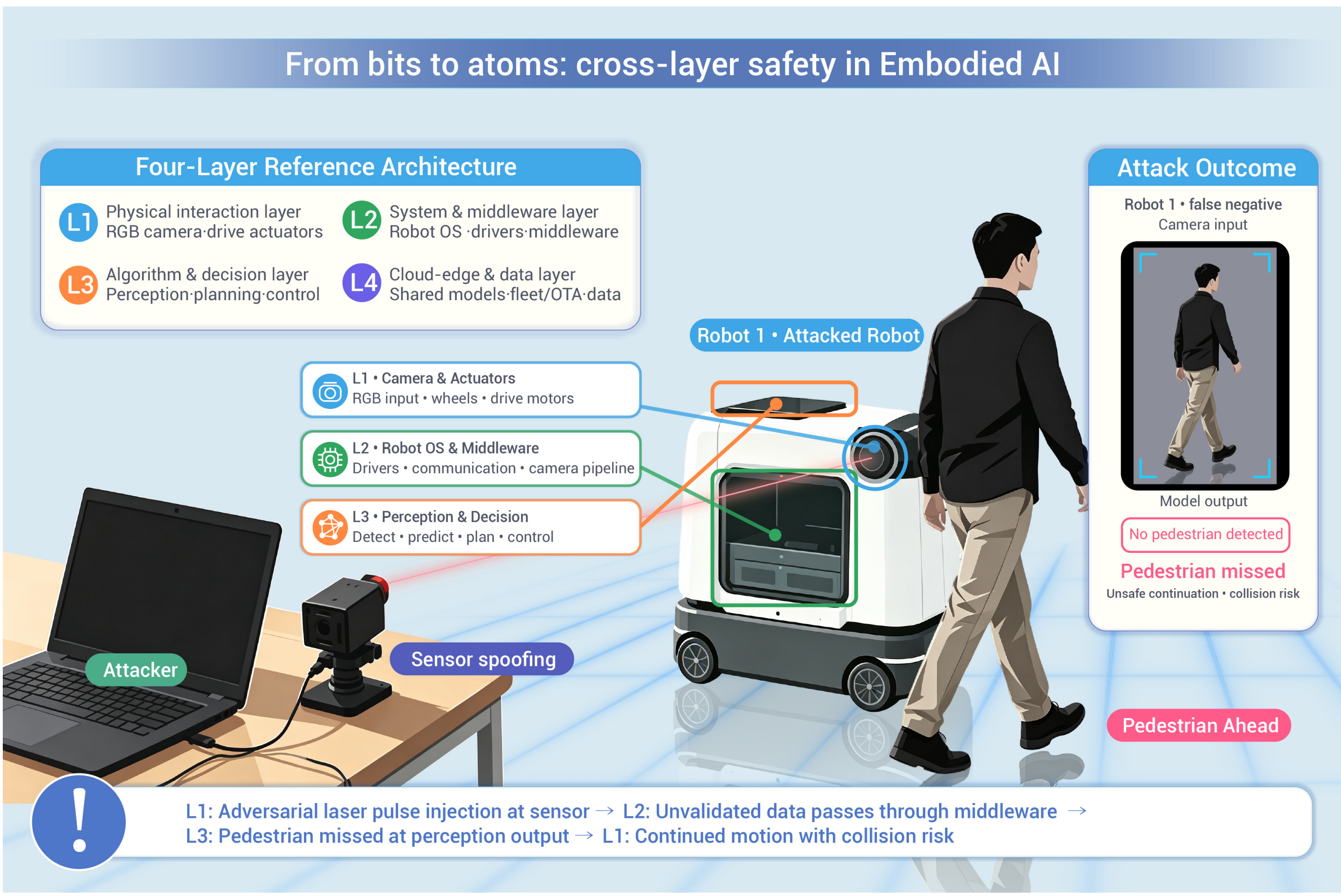

## PUBLIC SUMMARY

- Embodied Artificial Intelligence (AI) connects digital intelligence with physical action, so local faults, cyberattacks, or unsafe decisions can cause real-world harm.
- This review presents a four-layer framework spanning physical interaction, system and middleware, algorithm and decision, and cloud-edge and data.
- It reveals how risks propagate across layers and unifies functional, physical, and behavioral safety with defenses and closed-loop benchmarks.

# From bits to atoms: A survey of cross-layer safety in Embodied AI

Tuo Feng,[1] Yue Zhang,[2,*] Shiji Zhou,[3,*] Zhaoxin Fan,[3,*] Wenguan Wang,[1,*] Xiaowei Chi,[4] Tianyu Shen,[5] Xiangtai Li,[6] Xuesong Bai,[3,7] Zheng Zhu,[8] Sirui Han,[4] Boyu Zhou,[9,10] Zhaosong Wang,[11] Shizhe Chen,[12] Zilong Zheng,[13] Chenjia Bai,[14] Jiangmiao Pang,[15] Yaodong Yang,[16] Hao Zhao,[17] Di Hu,[18] Li Yi,[17] Yan Huang,[19,20] Qi Wu,[21] Hanwang Zhang,[6] Fei Gao,[1,10] Shanghang Zhang,[16] Hongsheng Li,[22] Jianye Hao,[23] Yuan Zhang,[24] Yingjie Xia,[25] Jian Shen,[26] Cewu Lu,[27] Hui Zhang,[28] Kai Chen,[29] Wenwu Yu,[30] Badong Chen,[31] Jungong Han,[17] and Yike Guo[4]

[1]Zhejiang University, Hangzhou 310027, China
[2]Shandong University, Jinan 250100, China
[3]Beihang University, Beijing 100191, China
[4]The Hong Kong University of Science and Technology, Hong Kong SAR 999077, China
[5]Beijing University of Chemical Technology, Beijing 100029, China
[6]Nanyang Technological University, Singapore 639798, Singapore
[7]State Key Lab of Intelligent Transportation System, Beijing 100088, China
[8]GigaAI, Beijing 100080, China
[9]Southern University of Science and Technology, Shenzhen 518055, China
[10]Differential Robotics (Hangzhou) Technology, Hangzhou 311100, China
[11]Shanghai CHOHO Robotics Co., Ltd., Shanghai 201315, China
[12]Inria, École normale supérieure, CNRS, PSL Research University, Paris 75013, France
[13]Beijing Institute for General Artificial Intelligence, Beijing 100080, China
[14]Institute of Artificial Intelligence of China Telecom (TeleAI), Shanghai 200030, China
[15]Shanghai Artificial Intelligence Laboratory, Shanghai 200232, China
[16]Peking University, Beijing 100871, China
[17]Tsinghua University, Beijing 100084, China
[18]Renmin University of China, Beijing 100872, China
[19]Institute of Automation, Chinese Academy of Sciences, Beijing 100190, China
[20]FiveAges, Hangzhou 310030, China
[21]Adelaide University, Adelaide, SA 5005, Australia
[22]The Chinese University of Hong Kong, Hong Kong SAR 999077, China
[23]Tianjin University, Tianjin 300072, China
[24]Fudan University, Shanghai 200433, China
[25]Hangzhou Dianzi University, Hangzhou 310018, China
[26]Zhejiang Sci-Tech University, Hangzhou 310018, China
[27]Shanghai Jiao Tong University, Shanghai 200240, China
[28]Hunan University, Changsha 410082, China
[29]Institute of Information Engineering, Chinese Academy of Sciences, Beijing 100085, China
[30]Southeast University, Nanjing 210096 China
[31]Xi'an Jiaotong University, Xi'an 710049, China
*Correspondence: zyueinfosec@sdu.edu.cn (Y.Z.); zhoushiji25@buaa.edu.cn (S.Z.); zhaoxinf@buaa.edu.cn (Z.F.); wenguanwang@zju.edu.cn (W.W.)

Embodied Artificial Intelligence (AI) introduces cyber–physical risks because digital failures and attacks can cause physical harm. This survey organizes embodied AI safety through a four-layer reference architecture — comprising the physical interaction, system and middleware, algorithm and decision, and cloud–edge and data layers — and uses it to connect vulnerabilities, propagation paths, and safeguards across the system stack. It traces the evolution of safety paradigms and examines how faults and attacks cross architectural boundaries to produce physical consequences. Furthermore, the survey systematizes composite attack chains, outlines multi-layered defense strategies, and reviews representative embodied safety benchmarks. Ultimately, we advocate for unifying software alignment with rigorous hardware constraints to build trustworthy systems capable of operating safely within human-centric environments.

## INTRODUCTION

The advancement of artificial intelligence has established Embodied AI as a transformative technology, marking a paradigm shift from virtual algorithmic operations to physical interactions.[1,2] Defined by autonomy, embodiment, and cognition,[3] these systems – ranging from service and industrial robots to drones and autonomous vehicles – expand the threat surface because digital vulnerabilities can propagate through perception, decision-making, and control to cause physical harm. As Embodied AI moves toward deployment, industry focus remains skewed toward algorithmic precision and cost optimization. Security mechanisms are relegated to secondary concerns or implemented as afterthoughts. This can leave interconnected components insufficiently protected, creating security gaps that threaten the safe and reliable operation of embodied platforms.[4,5]

This situation is exacerbated by several limitations. First, conventional safety-alignment methods developed for disembodied Large Language Models (LLMs) focus primarily on textual policy alignment while providing limited coverage of physical-world constraints. For disembodied LLMs, safety relies on textual alignment to filter toxic outputs;[6,7] however, text-based guardrails alone do not verify whether a generated action satisfies kinematic, environmental, or hardware constraints. An instruction containing no explicitly harmful language may induce hazardous physical actions in complex environments, because semantic safety checks alone cannot fully capture physical-world risks.[8–10] Second, hardware constraints and environmental dynamics need to be considered in high-level decision-making as well as low-level execution safeguards.[3] When these mechanisms are insufficiently coordinated, attacks on perception or localization may propagate into unsafe actions without being detected by high-level semantic checks.[11] For example, LiDAR spoofing can impair obstacle detection,[12] while GPS spoofing can corrupt or disrupt localization.[11] If these failures are not detected and contained, they may lead to unsafe navigation, collisions, or other physical harm. Third, assessing cybersecurity and functional safety in isolation can overlook risks arising from interactions among software, sensors, and actuators.[3,13] Compromised control software can alter physical behavior, while physical-layer sensor spoofing can corrupt perception and mislead downstream decisions, illustrating cyber-to-physical and physical-to-cyber risk propagation.[3,14] Joint analysis of cyber and physical dependencies is therefore needed to identify multistep attack paths and assess their potential system-level consequences.[15]

Safe deployment of embodied AI calls for an integrated approach to cybersecurity and safety. Localized faults and attacks can propagate through data dependencies and closed-loop interactions, potentially compromising system-level safety. Safety and security should therefore be integrated into system architecture, with coordinated monitoring and intervention across the cyber-physical loop.[16,17] Specifically, we advocate complementing instruction-following alignment with "action-feasibility alignment," checking proposed plans and actions against physical constraints while retaining semantic and contextual safety requirements. To address risks in dynamic environments, we advocate runtime assurance mechanisms and hardware safeguards appropriate to the platform and its risks, designed to block unsafe commands and mitigate hazards during execution. These mechanisms should be evaluated through interactive, closed-loop benchmarks that assess multimodal and cross-layer risks. By unifying this software alignment with rigorous hardware constraints and closing the software–hardware coupling gap, we can build trustworthy systems capable of operating safely within human-centric environments.[18,19] Motivated by this critical security vacuum, this survey aims to bridge the persistent gap in software–hardware coupled research. We systematically construct a comprehensive defense blueprint tailored to the interactive nature of embodied systems. The significance of this work lies in revealing how risks propagate from digital bits to physical atoms, proposing a holistic security paradigm that transcends traditional textual alignment and purely mechanical redundancy to realize behaviorally safe and physically trustworthy systems.

The remainder of this paper follows a progressive logic. "Background" section introduces a four-layer reference architecture and traces the evolution of embodied behavioral safety. "Hardware Safety and Security in Embodied AI" section analyzes the physical defense line, exploring hardware vulnerabilities across sensors, chips, and actuators. "Software Safety and Security in Embodied AI" section dissects the software and data layers, covering perception algorithms, LLM agents, and system middleware. "Compound Safety and Security Risks under Software–Hardware Coupling in Embodied AI" section delves into software–hardware coupling to reveal compound attack chains and propose cross-layer defenses. "Embodied Security and Safety Benchmarks" section surveys the evolution of safety evaluation frameworks and summarizes mainstream embodied safety benchmarks. Finally, "Future Directions and Outlook" section outlines future research directions, such as autonomous security evolution and unified world models, before "Conclusion" section concludes the survey.

## BACKGROUND

This section introduces a four-layer reference architecture, clarifies the safety concepts adopted in this survey, and traces the cumulative evolution of functional, physical, and behavioral safety. Building on these foundations, it examines cross-layer risk propagation and introduces the Identification–Mitigation–Evaluation (IME) organizing principle. Finally, it positions the survey within the existing literature and defines its scope.

### A four-layer reference architecture for embodied systems

Embodied AI differs from disembodied AI in one operationally important respect: its closed loop includes physical actuation and environmental feedback, meaning that failures in perception, planning, or control can be amplified by physical dynamics before correction.[1,2] A purely modular, component-level analysis is therefore insufficient for reasoning about safety, because the same cyber artifact (*e.g.*, a perception model) can be benign in a disembodied setting and hazardous once coupled with actuators. A layered view that tracks how information and energy flow between the digital and physical domains therefore provides a systematic basis for safety analysis.

Layered reference architectures for robotic systems have a long lineage. Early proposals like the NIST 4D/RCS model decomposed unmanned-vehicle control into sensing, world modeling, value judgment, and behavior generation.[20] Subsequent software-engineering studies systematized the evolution from object-oriented to component-based, service-oriented, and finally cloud-native robotic architectures,[21,22] while the cloud-robotics literature incorporated network services for shared learning and big-data processing.[23,24] Standardization bodies have codified much of this consensus: the IEEE 1872 family provides a shared ontology across robotics and automation, and ISO 22166 formalizes modularity for service robots at the component and interface level. Building on these foundations, we adopt a four-layer reference architecture as a platform-independent analytical abstraction that integrates the classical physical-control stack with the modern foundation-model stack.

Layer 1 (Physical Interaction Layer) comprises the sensors, actuators, power supply, and low-level communication that constitute the direct physical interface between robot and environment. This layer inherits the concerns of physical human–robot interaction[25] and is the last line of defense against physical harm. Layer 2 (System and Middleware Layer) encompasses the real-time operating system, drivers, communication buses, and middleware such as ROS and DDS that mediate between hardware and higher-level algorithms.[26,27] Layer 3 (Algorithm and Decision Layer) captures perception fusion, planning, control, and the increasingly prominent foundation model agents. The rapid progress of language-grounded and vision-language-action (VLA) models[28–33] has reshaped this layer from task-specific policies into an integrated cognitive substrate, with end-to-end models now driving high-level behavior. Layer 4 (Cloud–Edge and Data Layer) covers over-the-air update pipelines, cloud-side training infrastructure, cross-embodiment datasets, and privacy-sensitive data flows connecting a fleet of robots to shared foundation models.[23,34] This abstraction also applies to connected autonomous-driving platforms: cameras, LiDAR, radar, GNSS/IMU, and steering, braking, and propulsion actuators constitute L1; in-vehicle operating systems, device drivers, communication stacks, and middleware constitute L2; localization, perception, prediction, planning, and motion control constitute L3; and cloud-hosted HD maps and models, fleet data, remote diagnostics, and OTA services constitute L4.[35,36] The layers interact through identifiable functional interfaces, although these are logical rather than universally fixed physical boundaries. The L1/L2 boundary comprises device drivers and low-level buses or signals (*e.g.*, CAN and PWM); the L2/L3 boundary is mediated by middleware communication interfaces (*e.g.*, ROS 2/DDS topics); and the L3/L4 boundary comprises cloud–edge networking, remote APIs, and update or telemetry pipelines (*e.g.*, gRPC). Figure 1 gives a unified view of the four layers and their principal cross-layer data flows.

The four-layer architecture is an analytical reference, not the only decomposition of embodied AI. Its layers separate responsibilities and trust domains, expose characteristic failures and defenses, and provide interfaces for tracing risk propagation. Alternative formulations have distinct uses: perception–planning–control pipelines describe task flow but omit cloud and data lifecycles; cyber–physical models capture interactions but may merge middleware with learned decisions;[4,37] network models emphasize communication; and component stacks offer platform-specific detail. Thus, our architecture is not universally superior; it balances cross-layer traceability with platform independence and serves as our analytical coordinate system.

The four-layer decomposition is not merely a didactic device; it is the scaffold on which the rest of the survey is built. Each layer has characteristic failure modes that we will examine in subsequent sections, but the central observation is that these failures rarely remain confined to their layer of origin. Before that observation can be made precise, we must first tighten the notion of safety itself, which has undergone a three-stage paradigm shift

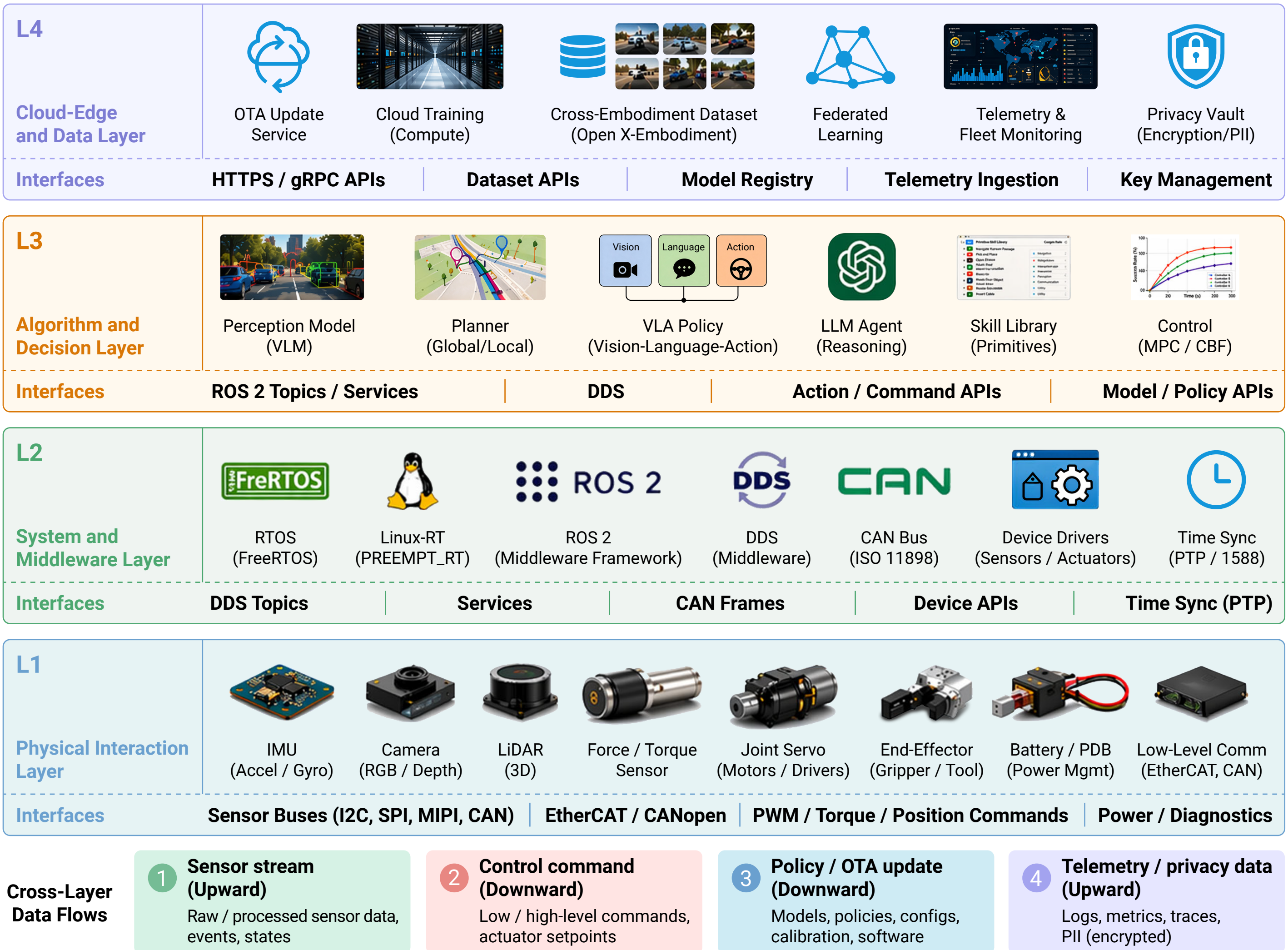

**Figure 1. Four-layer reference architecture for embodied AI systems** The stack progresses from physical interaction at the bottom (L1) to cloud–edge and data services at the top (L4). Representative components, technology stacks, and canonical interfaces are annotated within each layer. Four vertical arrows capture the principal cross-layer data flows that serve as risk-propagation channels. Abbreviations: OTA, over-the-air; PII, personally identifiable information; gRPC, gRPC Remote Procedure Calls; VLM, vision–language model; VLA, vision–language–action; LLM, large language model; MPC, model predictive control; CBF, control barrier function; ROS, Robot Operating System; DDS, Data Distribution Service; RTOS, real-time operating system; CAN, Controller Area Network; ISO, International Organization for Standardization; PTP, Precision Time Protocol; IMU, inertial measurement unit; LiDAR, light detection and ranging; PDB, power distribution board; I2C, inter-integrated circuit; SPI, Serial Peripheral Interface; MIPI, Mobile Industry Processor Interface (historical expansion); EtherCAT, Ethernet for Control Automation Technology; PWM, pulse-width modulation.

over the past three decades. That shift is the subject of the next subsection.

### Safety definition and paradigm evolution

The terms safety and security are often conflated in robotics and AI discourse, and the Chinese translation anquan covers both. A clean separation traditionally holds: safety addresses random faults and hazards arising from stochastic failures and unintended interactions, whereas security addresses deliberate adversarial actions. In embodied AI, however, their operational consequences often converge. A LiDAR spoofing attack is simultaneously a security incident (it is adversarial) and a safety incident (it can cause a physical collision), and the same perceptual failure mode can be triggered by either an adversarial laser pulse or an unfortunate reflection.[4,16] Recent standardization efforts, notably ISO/IEC TR 5469:2024 on functional safety and AI systems,[38] formalize this convergence and mark a decisive departure from the historical separation of the two fields.

Robotics and embodied AI overlap but are not synonymous. Robotics denotes the broader engineering field concerned with the design, control, and operation of physical robots, encompassing fixed, preprogrammed, teleoperated, and learned autonomous systems. Embodied AI refers more specifically to agents whose learned intelligence is grounded through a closed perception–reasoning–action loop involving interaction with an environment.[1,2] Thus, not every robot is an embodied AI agent, and embodied AI may also be studied in simulation without a physical robot. This survey focuses on physical embodied AI systems: robotic platforms that continuously couple learned perception, reasoning or planning, physical action, and environmental feedback. Accordingly, we use "robot" or "robotic system" when referring to the physical platform, conventional control mechanisms, or robotics standards, and "embodied AI system" or "embodied agent" when referring to the integrated learned system and its cross-layer or behavioral risks.

We adopt a unified definition: safety in embodied AI is the system's ability to remain controllable, interpretable, recoverable, and aligned under the joint presence of random faults, adversarial attacks, environmental anomalies, and model misspecification. These properties are grounded in distinct research traditions. Controllability bounds the system's action space in all reachable states, formalized through control barrier functions[39] and shielded reinforcement learning.[40] Interpretability requires that decision reasoning be intelligible to humans, drawing upon the paradigms in explainable agents[41] and explainable robotics.[42] Recoverability is the capacity to detect, isolate, and compensate for faults at runtime, drawing on classical fault-tolerant robotics[43] and runtime verification.[44] Alignment refers to the coherence between the system's emergent behavior and human intent, central to the AI-alignment literature.[45,46] To operationalize this definition, each attribute can be evaluated through concrete criteria: controllability is measured by the satisfaction rate of control barrier functions (CBFs) alongside the response latency of hardware safety interlocks; interpretability is quantified by the traceability of action outputs back to their multimodal inputs; recoverability is assessed by the success rate and execution time of autonomous fallback routines following fault injections; and alignment is evaluated based on task completion rates under constrained, human-centric reward models. Our attributes overlap substantially with the RICE framework (Robustness–Interpretability–Controllability–Ethicality),[46] but differ in two respects. First, we treat recoverability independently from robustness, because physical actions in the embodied setting are typically irreversible and thus require mechanisms for post-failure compensation that are qualitatively different from pre-failure robustness. Second, our notion of alignment is broader than ethicality and explicitly includes intent alignment over long-horizon tasks, a concern negligible for static classifiers but dominant for embodied agents operating over minutes or hours.

Against this definition, the history of robotic safety reflects a three-stage paradigm evolution, traced in Figure 2 and Table 1. The functional-safety paradigm (IEC 61508[47] and ISO 10218[48]) treats safety as the prevention of harm from random hardware failures and systematic software faults. Its methodological core is hazard and operability analysis and the allocation of Safety Integrity Levels (SIL) to safety-related functions. Work in this tradition, notably the STAMP/STPA framework,[49] focuses on Layers 1 and 2 and has matured into a highly codified engineering discipline. The physical-safety paradigm emerged with collaborative robotics. Codified in ISO/TS 15066,[50] it defines collaboration modes and tabulates human biomechanical limits for force, pressure, and energy. The seminal work of Haddadin *et al.*[51] provides the empirical grounding for these limits, and a rich literature on safe human–robot interaction has grown around them.[18,52] This paradigm couples Layers 1 and 3 through compliance control and contact detection. The behavioral-safety paradigm is the most recent and least codified. Triggered by autonomous vehicles, and now acutely relevant to large-model-driven robots, it concerns hazards that arise not from component failure but from the agent's own decisions in the face of an open, partially observed world. Its standards footprint is still forming – UL 4600,[53] ISO 21448 SOTIF, and ISO/IEC TR 5469:2024[38] are principal touchstones – and its empirical base is growing rapidly through work on jailbreak attacks on embodied language models,[54] safety benchmarks for embodied LLM agents,[10] and sandboxed risk identification for tool-using agents.[55]

Importantly, these three paradigms are cumulative rather than substitutive. An embodied AI system still requires mechanical interlocks (functional safety) and compliant contact (physical safety); behavioral safety is an additional layer of concern, not a replacement. Traditional robotic safety therefore remains within the scope of this survey because it provides the hardware, control, and middleware foundations on which physical embodied AI systems operate. This cumulative view motivates our treatment of safety as a cross-cutting property over all four architectural layers rather than a property of any single layer. These attributes serve as operational objectives herein: threats are analyzed by the capabilities they degrade, and defenses by those they restore. For instance, sensor spoofing may simultaneously degrade controllability, interpretability, and recoverability.

### Cross-layer risk propagation and survey organization

Having established the architectural coordinates and the safety semantics, we now turn to the central empirical observation that motivates this survey: risks in embodied AI rarely remain confined to their originating layer. The cyber-physical systems (CPS) literature established this point at the infrastructure scale,[4,16,37,56] and work on cascading failures in interdependent networks[57] provides an analogical framing from statistical physics: couplings between subsystems transform localized faults into system-level failures. The same logic applies to embodied AI, where the coupling is even tighter because a single closed loop spans all four architectural layers.

Three canonical propagation chains will recur throughout this survey and together establish the need for a cross-layer treatment. Chain A (physical-sensor spoofing) originates in Layer 1, where adversarial optical,[58] laser,[12] or acoustic[59] signals are injected into sensors, as illustrated by an adversarial laser pulse injected at the sensor (A1). The corrupted raw data propagate upward to Layer 3, where perception models produce confidently wrong outputs, such as a ghost object materialized in the perception output (A2). Planners then convert these outputs into unsafe control commands that return to Layer 1 through the actuators, as illustrated by an emergency brake triggered by a phantom obstacle (A3). Chain B (middleware-level compromise) originates in Layer 2, where controller-level vulnerabilities demonstrated by Quarta *et al.*[60] or targeted communication-channel attacks on teleoperated systems allow an adversary to inject or modify control commands, bypassing the algorithmic stack entirely. Figure 3 illustrates this path through an unauthenticated publisher joining the DDS domain (B1), the execution of a spoofed velocity setpoint (B2), and the actuator consequently exceeding its safe operating envelope (B3). Chain C (data-side contamination) originates in Layer 4, where a poisoned training corpus or a compromised OTA channel embeds a backdoor into the policy weights. The representative path in Figure 3 begins with a poisoned trajectory merged into the training set (C1), resulting in a backdoor embedded in the VLA policy weights (C2). During deployment, indirect prompt injection or a carefully crafted observation may activate the dormant policy, as illustrated by activation through a trigger scene (C3), producing unsafe physical actions far removed in both time and space from the original compromise. These three chains correspond to hardware, software, and data origins, and collectively establish that cross-layer propagation is not an incidental phenomenon but a systemic feature of embodied AI. The A1–A3, B1–B3, and C1–C3 identifiers correspond directly to the successive failure states shown in Figure 3. It is important to note that these three chains are illustrative rather than taxonomic. They serve as representative examples to demonstrate the systemic nature of cross-layer risks, rather than providing an exhaustive classification of all possible inter-layer propagation modes. Developing a comprehensive formal taxonomy of cross-layer vulnerabilities remains a critical open problem for future research.

To organize this complexity, we adopt an Identification–Mitigation–Evaluation (IME) analytical triad that structures the safety discussion throughout the survey. This triad is not an invention of this work; it is a specialization of well-established methodological frameworks. The identification step inherits from classical threat modeling, including attack trees,[61] STRIDE,[62] PASTA,[63] and their extensions to connected and autonomous vehicles via ISO/SAE 21434.[64] The mitigation step aggregates defensive mechanisms ranging from hardware redundancy and secure boot at Layer 1, through authenticated middleware at Layer 2, to safe reinforcement learning and runtime shielding at Layer 3, and differential privacy and secure update pipelines at Layer 4. The evaluation step builds on the physics-based detection literature[37] and on the recent wave of AI-agent safety benchmarks.[10,55] The NIST Cybersecurity Framework 2.0[65] and the NIST AI Risk Management Framework[66] provide particularly close precedents: their Identify–Map and Measure functions map cleanly onto our identification and evaluation phases, while Protect/Respond/Recover and Manage correspond to mitigation.

The four architectural layers map onto the survey sections not one-to-one but by coupling strength. Layer 1 possesses a high degree of physical independence and constitutes the last tangible line of defense between the system and the world; it is addressed bottom-up in the "Hardware Safety and Security in Embodied AI" section. Layers 2, 3, and 4 exhibit strong logical and data-flow coupling—middleware services are consumed by algorithms, algorithms are trained and updated through cloud–edge pipelines, and data flows pass through middleware—so we integrate them into the "Software Safety and Security in Embodied AI" section. The cross-layer propagation chains introduced above cannot be treated within any single-layer section and are therefore consolidated into the "Compound Safety and Security Risks under Software–Hardware Coupling in Embodied AI" section. Finally, evaluation is a cross-cutting concern that applies uniformly to all layers and all safety properties, and is addressed in the "Embodied Security and Safety Benchmarks" section. This reorganization follows the same principle used by El-Rewini *et al.*[67] for vehicular safety, namely restructuring by coupling strength rather than by nominal layer count. Figure 3 visualizes the three representative propagation chains that motivate this coupling-based organization.

The IME triad and the coupling-based section mapping provide the structural logic of this survey. The next three sections thus jointly apply three coordinates: the four-layer architecture locates where risks propagate, the four attributes define what degrades, and the IME framework specifies how

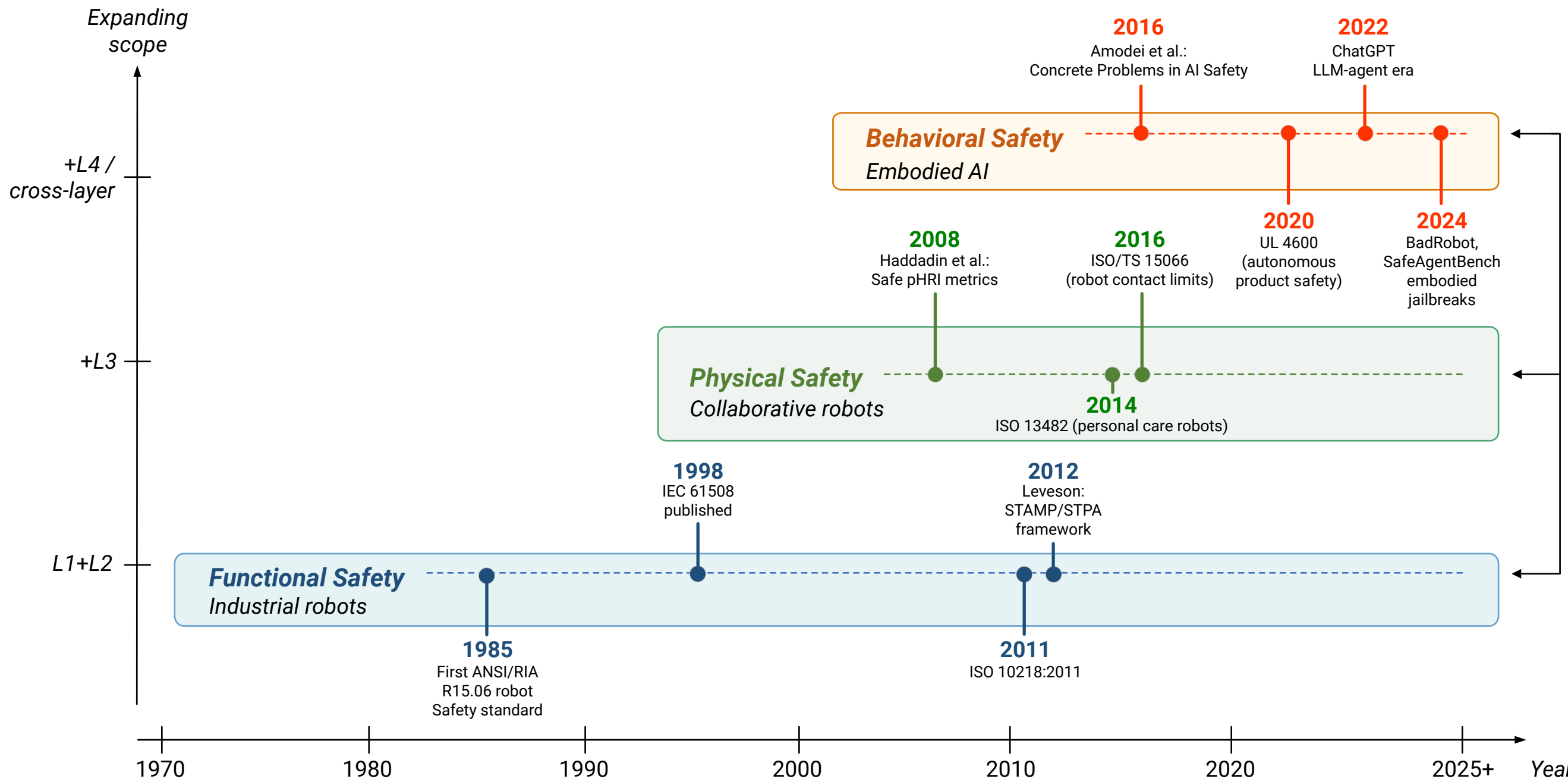

**Figure 2. Evolution of safety paradigms in robotic and embodied AI systems** Three horizontal bands, stacked by expanding scope of architectural coverage, trace the temporal emergence of functional safety (industrial robots), physical safety (collaborative robots), and behavioral safety (embodied AI). Each band extends to the right edge of the timeline, reinforcing that no paradigm has been superseded: all three remain active concerns today. Milestone markers on each band correspond to standards publications and landmark scholarly works discussed in the text. Abbreviations: GPT, generative pre-trained transformer (as in ChatGPT); ANSI, American National Standards Institute; RIA, Robotic Industries Association; IEC, International Electrotechnical Commission; TS, technical specification; UL, Underwriters Laboratories; pHRI, physical human–robot interaction; STAMP, System-Theoretic Accident Model and Processes; STPA, System-Theoretic Process Analysis.

Table 1. Evolution of safety paradigms in robotic and embodied AI systems.

| | Functional Safety *(Industrial robots)* | Physical Safety *(Collaborative robots)* | Behavioral Safety *(Embodied AI)* |
|---|---|---|---|
| Primary concern | Random and systematic component failures | Unsafe physical contact during proximity | Unsafe decisions in open, partially observed worlds |
| Layers emphasized | L1 + L2 | L1 + L3 | L1 + L2 + L3 + L4 |
| Representative standards | IEC 61508; ISO 10218 | ISO/TS 15066; ISO 13482 | UL 4600; ISO 21448; ISO/IEC TR 5469 |
| Core methods | Hazard analysis, SIL allocation, redundancy | Force/pressure limits, impedance control, safe-rated stop | Value alignment, runtime monitoring, safety cases |
| Canonical risks | Mechanical failure, sensor fault, control malfunction | Collision, clamping, excessive contact force | Jailbreak, specification gaming, cross-layer cascades |

The three paradigms are cumulative: each subsequent stage adds new concerns without removing those of previous stages. Section coverage is shown in Table 2.

to respond (detailed mapping in Supplement).

## Survey scope and comparison with related surveys

Safety and security in robotic and AI systems have been the subject of considerable survey literature, which we group into six strands. The first covers robot cybersecurity and human–robot interaction safety.[5,13,18,52,68,69] The second addresses cyber-physical systems security,[4,37] including control-theoretic extensions.[70] The third spans autonomous-vehicle safety, ranging from foundational attack-surface analyses[71] to comprehensive cybersecurity treatments.[35,36,67] The fourth explores physical adversarial attacks on perception,[72] with particular emphasis on autonomous driving.[73] The fifth and most rapidly growing strand examines LLM- and agent-level safety.[74–76] Finally, the sixth and most recent strand focuses directly on embodied AI safety, encompassing broad vulnerability reviews, navigation safety, and institutional governance frameworks.[3,77–79]

Table 2 presents a coverage matrix along seven dimensions: the four architectural layers (L1–L4), behavioral safety, compound cross-layer risks, and evaluation benchmarks. Four gaps emerge from this analysis. First, the cross-layer coverage gap: no existing survey rigorously addresses all four layers. Robot-cybersecurity surveys cover L1 and L2 but treat L3 only superficially; LLM-agent surveys are almost entirely L3-centric, ignoring embodied hardware and middleware; AV surveys cover L1, L2, and L4 but sidestep the L3 algorithm-decision layer as reshaped by foundation models; CPS surveys are strong on L1 and L2 but pre-date the large-model era at L3. Second, the behavioral-safety gap: traditional robot and CPS surveys rarely engage with value alignment or agent decision risks, while LLM-agent surveys discuss jailbreaks and prompt injection but treat their physical consequences only anecdotally. Third, the compound-risk gap: with the partial exceptions of a few works, such as Haskard and Herath,[68] existing surveys present threats as per-component or per-layer lists and do not develop a taxonomy of cross-layer compound risks. Fourth, the benchmarking gap: while some LLM-agent surveys enumerate text-level safety benchmarks, none systematically integrates embodied, cross-layer safety benchmarks such as SafeAgentBench,[10] RoboPAIR,[8] BadRobot,[54] POEX,[80] and SOTIF-oriented driving benchmarks.[81]

Against this backdrop, our survey makes five differentiating contributions. First, we provide a comprehensive and unified treatment of embodied AI safety along a four-layer architecture, bridging traditional robot safety, CPS security, AV security, LLM-agent safety, and VLA safety under a single taxon-omy. Second, we introduce behavioral safety as an explicit and independent dimension, covering value alignment, long-horizon planning risk, tool misuse, and multi-agent collusion, bridging the semantic gap between the LLM-agent and the robot-safety literatures. Third, we develop an analytical framework for representative cross-layer propagation patterns that identifies causal chains (*e.g.*, sensor spoofing → VLA jailbreak → physical harm, data poisoning → weight backdoor → field trigger), with literature-grounded worked examples reconstructed from published experiments. Fourth, we systematically integrate embodied safety benchmarks across the four layers and the behavioral and compound-risk dimensions, filling a persistent gap in existing surveys. Fifth, we argue for embodied AI safety as an independent subfield by showing that it is not reducible to the union of LLM-agent safety and robot safety: the embodiment introduces semantic–physical misalignment, irreversible action effects, state–action gaps, and compositional safety gaps that neither parent discipline can handle in isolation.

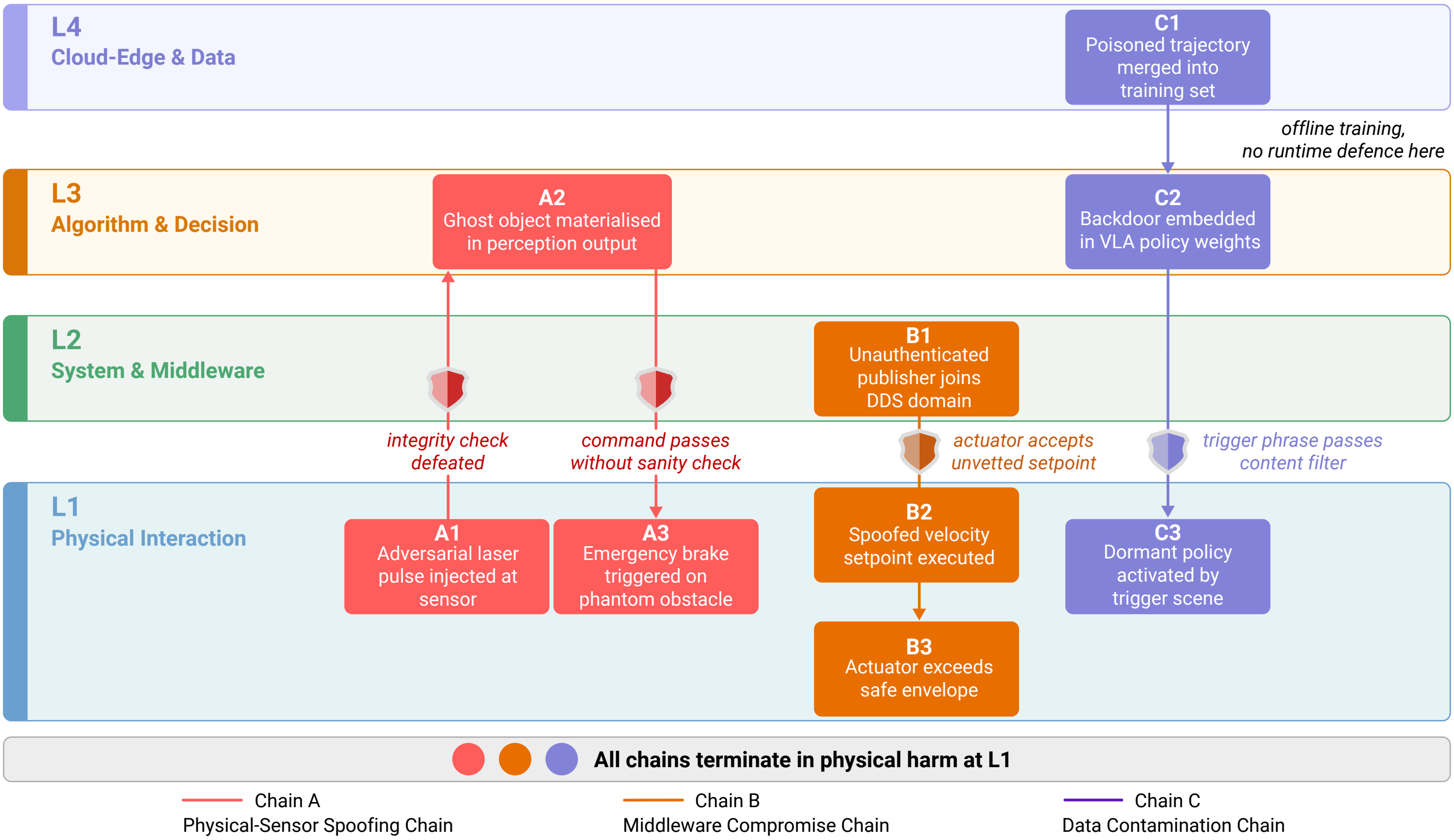

**Figure 3. Cross-layer risk propagation in embodied AI** The figure highlights three representative failure chains: Chain A (physical–sensor spoofing), Chain B (middleware compromise), and Chain C (data contamination). The nodes A1–A3, B1–B3, and C1–C3 mark successive failure states that propagate from Layers 1, 2, and 4, respectively; all three chains ultimately terminate in physical harm at the physical-interaction layer.

We close by defining what lies outside our scope. We do not treat content safety of disembodied language models, general AI-alignment theory, non-embodied robotic teleoperation security, or the mechanical-reliability engineering of robot hardware. Nor do we address the broader societal, legal, and economic implications of deploying embodied AI, which are the subject of a distinct policy literature. Although this survey does not address legal or economic policy, its four-layer framework supports governance by making system responsibilities, failure propagation, and verifiable safeguards explicit across the deployment lifecycle. These technical mappings can inform risk-based certification, auditing, accountability, and public trust. Within these boundaries, the next four sections develop the four-layer, four-attribute, cross-layer, and evaluation-benchmark treatment promised by this background.

## HARDWARE SAFETY AND SECURITY IN EMBODIED AI

Embodied AI is fundamentally distinct from disembodied algorithmic systems: its operational closed loop inherently couples digital decision-making with continuous physical actuation and environmental feedback. The move from virtual to physical deployment redefines security priorities: hardware-level security becomes the foundation of a trustworthy embodied system.[4] Consistent with the unified definition in this survey, hardware abnormalities may have non-adversarial or adversarial origins. Environmental disturbances, calibration drift, intermittent power, thermal stress, component wear, actuator backlash, and random hardware faults may degrade sensing or control without malicious intervention. Deliberate attacks, including sensor spoofing, fault injection, firmware manipulation, and actuator-command tampering, may intentionally produce similar corrupted states. Although their initiating causes differ, both may propagate from Layer 1 into perception, planning, and control and ultimately produce the same unsafe physical consequence. Figure 4 provides an overview of the hardware safety and security landscape discussed in this section.

### Sensor security

As the primary interface linking embodied agents to the physical world, sensors face pervasive cross-domain attack risks. In a VLA pipeline, spoofed camera, LiDAR, IMU, GPS, or tactile signals are encoded as corrupted multimodal representations or observation tokens, potentially creating false objects or affordances, concealing hazards, and altering subsequent reasoning, planning, and action generation.

***Visual sensor spoofing.*** Vision sensors are currently the primary pathway for multimodal embodied AI to acquire rich environmental information for tasks like SLAM, object recognition, and navigation.[82] However, because these sensors convert optical signals into digital inputs, they are highly susceptible to adversarial manipulations.[83] Within the realm of physical and cross-domain threats, key adversarial techniques include adversarial patches,[84] light projection attacks,[85] exploiting the light spectrum,[86] and cross-domain acoustic attacks.[87] These perception-targeted attacks can severely disrupt real-time decision-making in autonomous systems,[88] leading to dangerous consequences such as: causing traffic-light color misclassification (*e.g.*, red lights as green);[89] manipulating object detection to fabricate/erase obstacles;[90] tampering with lane detection via crafted road markings to force vehicles off-path;[91] disrupting multi-object tracking by altering trajec-

Table 2. Coverage matrix of representative related surveys.

| Survey | Year | Venue | L1 | L2 | L3 | L4 | Behav. | Comp. | Bench. | Primary focus |
|---|---|---|---|---|---|---|---|---|---|---|
| *Robot cybersecurity & HRI safety* | | | | | | | | | | |
| Lasota *et al.* [18] | 2017 | FnT Robotics | ✓ | — | ○ | — | ○ | — | — | Safe HRI methods |
| Villani *et al.* [52] | 2018 | Mechatronics | ✓ | — | ○ | — | ○ | — | — | Industrial HRC safety |
| Zacharaki *et al.* [69] | 2020 | Safety Sci. | ✓ | — | ○ | — | ○ | — | — | HRI safety bounds |
| Yaacoub *et al.* [5] | 2022 | IJIS | ✓ | ✓ | ○ | ○ | — | ○ | — | Robot cybersecurity |
| Botta *et al.* [13] | 2023 | Intell. Syst. Appl. | ✓ | ✓ | ○ | ○ | — | ○ | — | Robot cybersecurity |
| Haskard and Herath[68] | 2025 | ACM CSUR | ✓ | ✓ | ○ | ○ | ○ | ✓ | — | Safety--trust--cyber nexus |
| *Cyber-physical systems security* | | | | | | | | | | |
| Humayed *et al.* [4] | 2017 | IEEE IoT J | ✓ | ✓ | ○ | ○ | — | ✓ | — | CPS unified framework |
| Giraldo *et al.* [37] | 2018 | ACM CSUR | ✓ | ○ | ○ | — | — | ✓ | ○ | Physics-based detection |
| Liu *et al.* [16] | 2019 | Acta Autom.\ Sin. | ✓ | ✓ | ○ | — | — | ✓ | — | Integrated CPS safety |
| *Autonomous-vehicle safety* | | | | | | | | | | |
| Petit & Shladover[71] | 2015 | IEEE T-ITS | ✓ | ✓ | ○ | ○ | — | ○ | — | AV attack surface |
| El-Rewini *et al.* [67] | 2020 | Veh.\ Commun. | ✓ | ✓ | — | ○ | — | ○ | — | Vehicular comm. security |
| Kim *et al.* [35] | 2021 | Comput.\ Secur. | ✓ | ✓ | ○ | ○ | — | ○ | — | AV cyber-attacks |
| Sun, Yu, Zhang[36] | 2022 | IEEE T-ITS | ✓ | ✓ | ○ | ✓ | — | ○ | — | CAV cybersecurity |
| *Physical adversarial perception attacks* | | | | | | | | | | |
| Guesmi *et al.* [72] | 2024 | IEEE Access | ✓ | — | ✓ | — | — | ○ | ○ | Camera-based phys.\ adv. |
| Girdhar *et al.* [73] | 2023 | IEEE OJ-VT | ✓ | ○ | ✓ | ○ | — | ○ | ○ | AV adversarial atk./def. |
| *LLM / AI agent safety* | | | | | | | | | | |
| Liu *et al.* [76] | 2023 | NeurIPS SoLaR Workshop | — | — | ✓ | — | ✓ | ○ | ✓ | LLM alignment evaluation |
| Deng *et al.* [74] | 2025 | ACM CSUR | — | — | ✓ | ○ | ✓ | ○ | — | AI agent threats |
| He *et al.* [75] | 2025 | ACM CSUR | — | — | ✓ | ✓ | ✓ | ○ | — | LLM agent sec./priv. |
| *Embodied AI (non-safety, for contrast)* | | | | | | | | | | |
| Duan *et al.* [1] | 2022 | IEEE TETCI | — | — | ✓ | — | — | — | ✓ | Embodied AI simulators |
| Liu *et al.* [2] | 2025 | IEEE/ASME T-Mech | ○ | ○ | ✓ | ○ | — | — | ○ | Embodied AI landscape |
| *Embodied AI safety (most directly related)* | | | | | | | | | | |
| Neupane *et al.* [77] | 2024 | IEEE Access | ✓ | ○ | ✓ | ○ | ○ | ✓ | — | AI-robotics security |
| Xing *et al.* [3] | 2026 | ACM CSUR | ✓ | ○ | ✓ | ○ | ✓ | ✓ | ○ | Embodied AI vuln./atk. |
| Xu *et al.* [79] | 2025 | Bull. CAS | ✓ | ✓ | ✓ | ✓ | ✓ | ✓ | ○ | EAI governance (Chinese) |
| Wang, Hu, Mu[78] | 2025 | IJCAI | ○ | — | ✓ | — | ✓ | ○ | ✓ | Embodied navigation safety |
| This survey | 2026 | — | ✓ | ✓ | ✓ | ✓ | ✓ | ✓ | ✓ | Four-layer unified treatment with behavioral safety as an explicit axis |

Symbols are operationally defined as follows: ✓ = systematic coverage supported by a dedicated section/subsection and structured synthesis; ○ = partial coverage (brief discussion, isolated examples, or lacking systematic synthesis); — = not covered. The scores indicate topical scope and depth rather than the overall quality of a survey. Behav. = behavioral safety; Comp. = compound cross-layer risks; Bench. = evaluation benchmarks. Entries are ordered chronologically within each strand.

tory predictions;[92] and manipulating drone velocity readings to induce instability and crashes.[93]

***LiDAR/depth sensor spoofing.*** LiDAR and depth sensors[94] rely on the reflection of emitted signals and Time-of-Flight (ToF) principles for 3D spatial mapping and obstacle detection.[95,96] However, these distinct operational mechanisms introduce severe hardware vulnerabilities. These sensors are highly susceptible to signal interference, environmental degradation, and spoofing attacks such as echo forgery, physical occlusion, or high-intensity malicious signal injection.[97] Specifically, attackers can exploit LiDAR through laser projection attacks: by synchronizing with the victim's LiDAR using a photodiode and manipulating the delay of laser pulses, adversaries can inject fake points into the sensor's point cloud. This allows them to fabricate nonexistent objects, distort the perceived positions of real obstacles, or completely hide actual hazards.[98] Such adversarial manipulations can cause autonomous systems and robotic platforms to perform abrupt stops or suffer fatal navigation errors.

***IMU/GPS spoofing.*** Inertial Measurement Units (IMUs) and GPS are the cornerstones of embodied AI navigation and positioning,[99] but both face severe threats from hardware spoofing attacks. IMUs are susceptible to sensor drift, noise, and environmental factors.[100,101] By emitting precisely modulated resonant acoustic signals, attackers can manipulate the internal mass-spring structures of the IMU to inject stable, forged signals into multi-axis inertial sensors.[102] This attack not only disrupts the target but also controls its linear or angular velocity,[103] causing embodied devices like drones to lose attitude control, deceiving object detection systems,[104] and even enabling malicious takeovers.[102] Similarly, GPS signals are easily obstructed in complex environments,[105] and the lack of robust signal authentication makes them highly vulnerable to jamming and spoofing.[106,107] Attackers can use GPS simulators or portable Software-Defined Radios (SDRs) to transmit counterfeit satellite signals with sufficient received power or synchronization to capture a vulnerable receiver.[108] After takeover, crafted pseudoranges or navigation messages can manipulate the receiver's estimated position, subject to receiver-tracking and sensor-fusion constraints.[109] Under demonstrated attack conditions, GPS or trajectory spoofing can cause autonomous vehicles to deviate from their intended paths,[110] corrupt LiDAR motion compensation so that safety-critical objects are missed or mislocalized,[111] and disrupt map-assisted traffic-light detection, producing red-light violations in end-to-end simulation.[112] If not detected and contained, these effects could lead to serious physical harm.

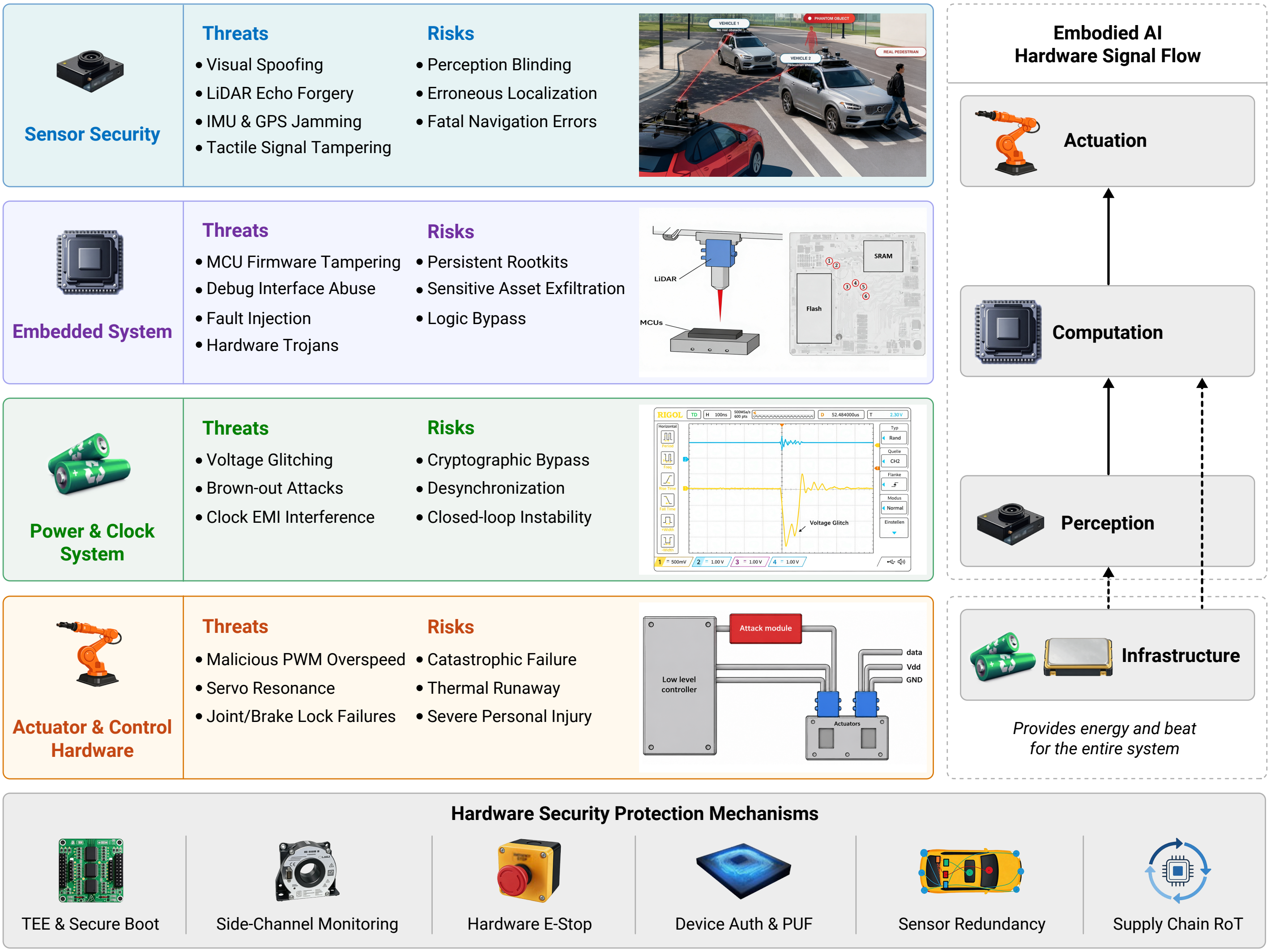

**Figure 4. Overview of hardware safety and security in Embodied AI** The figure illustrates the hardware-level failure and threat landscape along the physical signal flow: from perception, computation, and infrastructure, to actuation. It outlines the cross-layer vulnerabilities and establishes a hardware-enforced defense-in-depth architecture to support cyber-physical isolation and resilience. Abbreviations: GPS, Global Positioning System; MCU, microcontroller unit; EMI, electromagnetic interference; TEE, trusted execution environment; E-Stop, emergency stop; PUF, physically unclonable function; RoT, root of trust; SRAM, static random-access memory.

***Tactile and force sensor tampering.*** Tactile sensors measure force, torque, and pressure distribution, playing a vital role in precise manipulation tasks such as fragile object grasping, precision assembly, and human-robot collaboration (HRC).[113] Recent advancements have focused on miniaturization, soft stretchable materials, and multimodal integration to enhance robustness and versatility.[114,115] However, beyond natural physical wear, these embedded hardware components face severe security vulnerabilities. Manipulating sensor signals or compromising sensor communication interfaces can corrupt the feedback used by closed-loop controllers and consequently alter system actuation.[77,116] Furthermore, signal cross-talk between sensing elements in capacitive or iontronic sensor arrays can produce incorrect or misleading measurements of physical stimuli.[117] These adversarial manipulations lead to critical decision-making failures, resulting in crushed objects or severe personal injury.

## Chip and embedded system security

Across the compute spectrum – from edge nodes to microcontrollers – chips and embedded systems underpin both embodied large-model inference and low-level actuation.

***MCU/SoC firmware tampering.*** Microcontrollers (MCUs) and systems-on-chip (SoCs) implement core control functions, including sensor-data processing, actuator-command generation, and communication protocols. Without secure boot and authenticated firmware-update mechanisms anchored in a hardware root of trust (RoT), attackers who gain access to firmware storage or update interfaces may modify the bootloader or install malicious firmware that persists across reboots.[118] Such firmware compromises can interfere with sensor processing or actuator-command genera-tion, potentially causing unsafe behavior in embodied platforms.

***Debug interface abuse.*** Hardware debug interfaces, such as JTAG, UART, and SWD, are ubiquitous on device motherboards, conventionally provisioned to facilitate diagnostics, firmware flashing, and maintenance. However, if these interfaces remain enabled without appropriate access controls—such as lifecycle-based locking, readout protection, authenticated debug reopening, or permanent disablement—they can provide a physical attack path. Adversaries with physical access can use logic analyzers to inspect exposed signals or connect compatible debug probes to unsecured JTAG/SWD pins. When unrestricted JTAG/SWD debug access is available, a compatible debugger may halt or single-step the processor and access permitted memory regions. This privileged access allows malicious actors to exfiltrate highly sensitive assets or directly implant stealthy execution payloads. Furthermore, compromised debug interfaces enable attackers to silently modify critical actuator thresholds or bypass secure boot verifications entirely, allowing a low-level compromise to propagate into unsafe physical behavior.

***Side-channel attacks and fault injection (FI).*** Embodied AI devices, frequently exposed to open environments, are highly susceptible to physical-layer attacks. Side-channel attacks pose a significant threat by exploiting physical characteristics[119,120] to infer sensitive data like encryption keys during cryptographic operations, thereby compromising system confidentiality.[121] Furthermore, active fault injection poses a serious hardware-security threat. Attackers can use precise laser pulses to irradiate 32-bit MCUs, injecting transient faults during flash memory read operations.[122] This forcefully alters the read values and successfully bypasses underlying cryptographic counters or permission verification logic.

***Malicious component implantation and pre-installed backdoors.*** The prolonged, highly fragmented, and globalized nature of the hardware supply chain introduces profound vulnerabilities long before embodied AI systems are deployed.[123] Adversaries can insert malicious logic into third-party IP cores during design and integration or alter IC masks or layouts during fabrication.[124] Many such modifications are notoriously difficult to detect using functional testing, as they are designed to remain entirely dormant until activated by a highly specific sequence of physical triggers or environmental conditions.[125] Concurrently, malicious actors can pre-install logic bombs and dormant backdoors deep within the underlying system firmware or vendor-supplied board support packages (BSPs). Once triggered, these systemic threats can facilitate stealthy, long-term espionage, silently exfiltrate sensor telemetry, or execute synchronized, fleet-wide denial-of-service attacks that severely disrupt autonomous agents in the field.

### Power supply and clock system security

The reliable operation of high-performance computing systems and servo systems depends on stable power delivery and correct timing.

***Voltage glitching and power attacks.*** Adversaries can manipulate the supply voltage of an embodied device's SoC or MCU to inject hardware faults. By precisely timing transient voltage glitches, including brief undervolting events, attackers can induce setup and hold time violations in the underlying CMOS logic.[126] Such timing violations can cause computational or control-flow faults during cryptographic computations or authentication routines.[127] If analogous faults occur during VLA inference, they may corrupt intermediate computations or action outputs, causing the generated behavior to become inconsistent with the current observations and task instruction. If voltage-induced faults compromise code-authentication checks on an embodied platform, they may allow unauthorized software to be loaded and compromise the robot's control stack.

***Clock interference attacks.*** Modern embodied agents often use Real-Time Operating Systems (RTOSs) and hardware timers derived from local clock sources to schedule sensing, communication, and control tasks. Attackers can inject targeted Electromagnetic Interference (EMI) or physically tamper with clock pins to induce clock glitching—abruptly compressing or stretching the clock period. Such disturbances can corrupt data transfers or alter instruction execution, including instruction skips.[128] For foundation-model-driven agents, such disturbances may also desynchronize camera, LiDAR, and proprioceptive inputs, causing the VLA policy to reason over stale or inconsistent observations and generate unsafe action tokens or continuous control outputs. In physical robotic systems, such desynchronization can disrupt closed-loop control algorithms, including PID controllers, resulting in oscillatory or unstable motion that may endanger human operators.

### Actuator and control hardware security

Located at the terminal end of the signal flow, actuators are the critical components that translate the computational decisions of embodied large models into real-world physical actions. An actuator-level compromise may therefore modify or bypass an otherwise valid VLA command, creating a mismatch between model-generated actions and their physical execution. The resulting unexpected motion changes the environment and subsequent observations, allowing the original hardware anomaly to be amplified in the next VLA inference cycle.

***Motor drive anomalies and malicious control.*** Actuators convert electrical, hydraulic, or pneumatic energy into physical force and motion according to control commands. After compromising an embedded low-level controller, attackers may override VLA-derived commands and defeat safety constraints imposed by upstream planning modules. By manipulating PWM duty cycles, switching patterns, or current and velocity references while disabling drive-level limits, adversaries may cause excessive motor speed, torque, or current. Such low-level manipulation can defeat upstream collision-avoidance commands and result in collisions, unstable motion, or mechanical damage.

***Servo system overload, instability, and malfunction.*** Servo instability may arise without malicious intervention because of parameter drift, mechanical wear, thermal variation, lubrication degradation, or missed control deadlines. Similar unsafe behavior can also be induced deliberately by manipulating motion commands or controller parameters. After compromising low-level control, attackers can cause unexpected trajectories, oscillatory motion, or control instability.[129] Repeated aggressive actuation may additionally increase mechanical and thermal stress, with the resulting damage depending on the actuator design and its protection mechanisms.

***Brake, gripper, and joint lock failure.*** Brakes, grippers, and joint locks are safety-critical components that may fail without malicious intervention because of wear, wiring faults, power loss, or thermal stress. During hazardous conditions, electromechanical brakes and safety-rated emergency-stop circuits can serve as hardware-level safeguards against physical harm. Because these safety functions may depend on programmable or networked machine-control components, their control paths must also be protected against cyber manipulation.[130] Hardware man-in-the-middle devices placed on actuator communication lines can intercept or modify digital servomotor commands.[131] In systems where brakes or joint locks are electronically controlled, analogous manipulation of their control signals could prevent timely engagement or release. For example, unauthorized release of a joint lock while a robot is carrying a heavy payload, or unexpected brake activation during high-speed motion, may cause loss of load support, mechanical damage, or a dangerous payload drop.

### Hardware security protection mechanisms

In "Background" section, we introduced a four-layer reference architecture in which Layer 1 provides the physical foundation for all higher-layer functions. Consequently, hardware faults or attacks—from environmental degradation, calibration drift, power instability, and component wear to malicious firmware and physical tampering—may propagate through the entire system stack. We therefore organize the Layer 1 protection mechanisms reviewed below according to their defensive roles.

***Trusted execution environment (TEE) and secure boot.*** Hardware-enforced isolation provides an important foundation for computational security. A TEE implemented on the main SoC, such as ARM TrustZone, can isolate security-sensitive operations, including cryptographic key management, from the rich operating system (Rich OS).[132] Furthermore, a cryptographically authenticated secure boot process can establish a chain of trust for executable boot components. Starting from an immutable root of trust containing verification code and trusted key material, each boot stage verifies the integrity and authenticity of the next stage before transferring control, extending through the bootloader and operating system kernel.[133] VLA model weights, which are typically loaded after boot, require a separate integrity and authenticity check before use.

***Device authentication and remote attestation.*** In a decentralized multi-agent ecosystem, establishing the identity of each embodied entity is crucial. Physical Unclonable Functions (PUFs) exploit manufacturing-induced variations in integrated circuits to produce device-specific challenge–response behavior that can support device authentication and secret-key derivation.[134] PUF-derived secrets may therefore be used in challenge–response protocols to authenticate a device. Device authentication alone, however, does not establish platform integrity. When combined with trusted measurement and evidence-generation mechanisms, an authenticated device identity can bind attestation evidence to a particular platform. Semantic remote attestation can then allow a remote verifier to assess selected program properties before assigning critical tasks or releasing sensitive data.[135]

***Sensor redundancy and cross-validation.*** To counter physical-layer sensor attacks and interference (*e.g*, GPS jamming and LiDAR laser injection), embodied systems can reduce their reliance on any single modality through redundant sensing, heterogeneous sensor fusion, and cross-validation. Attack-resilient fusion can combine multiple measurements of the same physical variable while limiting the influence of compromised sensors, provided that sufficient trustworthy redundancy is available.[136] Demonstrated optical-flow spoofing attacks against UAVs further illustrate the risks of relying on a single visual estimator.[93] For example, if visual odometry indicates a rapid change in velocity or attitude while inertial measurements provide no corresponding acceleration or angular-velocity evidence, a consistency monitor can flag the discrepancy after accounting for sensor timing, noise, and uncertainty. The system may then reject the suspect measurement stream and enter a platform-appropriate safe fallback mode.

***Safety-related actuator control and emergency stop mechanisms.*** At the execution layer, safety-related stopping functions should be designed to remain effective even if the main application controller is compromised. For industrial robots, emergency-stop and protective-stop functions should not rely solely on ordinary application commands.[48] The appropriate stopping behavior and its implementation depend on the robot and its risk assessment. Protective devices and monitored safety limits can initiate a protective response, whereas emergency-stop devices provide a separate operator-initiated means of stopping hazardous motion. Software- and wireless-mediated safety interfaces can themselves become attack surfaces, underscoring the need to protect the complete stopping chain.[60]

***Abnormal power, temperature, and vibration monitoring.*** Electrical, thermal, and mechanical measurements can provide complementary indicators of abnormal operation. External power-consumption monitoring has been used to detect malware on embedded medical devices without modifying the devices.[137] On-chip thermal monitoring has also been investigated for detecting changes associated with activated hardware Trojans.[138] Vibration measurements near robot joints can help diagnose mechanical degradation, such as joint backlash.[139] Because these signals do not by themselves establish whether an anomaly is malicious, diagnostic alerts should be assessed alongside operating context; when a potential hazard is identified, an appropriately designed safety function can initiate a protective response.

***Supply chain tracking and RoT.*** Hardware Trojans may be introduced during integrated-circuit (IC) design or fabrication, making component assurance relevant before deployment.[124] A hardware root of trust (RoT) can anchor selected security functions, such as firmware-update authentication and integrity checking, but does not by itself establish that an IC or third-party IP core is free of malicious modifications.[118] To reduce counterfeit-component risk, developers can maintain component provenance across procurement and assembly, source parts from vetted suppliers, and apply risk-based inspection and testing.[140] These measures complement one another rather than provide a guarantee that all compromised components will be detected.

***Attribute and IME synthesis.*** Hardware faults and attacks can undermine controllability and recoverability: corrupted sensing, computation, timing, or actuation may drive the system outside its physical safety envelope or impair its ability to enter a safe fallback state. They can also weaken the traceability aspect of interpretability when actions cannot be reliably linked to sensor inputs, firmware state, or command sources, and affect alignment when executed actions diverge from human intent or safety requirements. At the hardware layer, risk identification combines signal-anomaly detection, integrity checks, and physical-state monitoring; mitigation uses threat-appropriate measures, including roots of trust for firmware integrity, sensor redundancy, fault containment, and safety-related interlocks. Evaluation should assess detection coverage, false-alarm rates, safety-constraint violations, intervention latency, and the success and timeliness of recovery to a safe state.

## SOFTWARE SAFETY AND SECURITY IN EMBODIED AI

Software safety and security are central to embodied AI because software-generated decisions can ultimately drive physical actuation. Unlike systems whose outputs remain in the digital domain, embodied AI couples perception, planning, control, middleware, and data flows with physical actuation and environmental feedback; human interaction is also part of many deployments.

Software faults and attacks can propagate across layers: erroneous or manipulated observations may mislead planners, unsafe plans may lead to hazardous commands if not intercepted by control-layer safeguards, and insufficiently protected middleware may allow unauthorized commands to reach control interfaces. Thus, software safety and security should be addressed at the system level rather than solely within isolated modules. Figure 5 provides an overview of the software safety and security landscape discussed in this section.

### Perception algorithm security

***Perception as the first security boundary.*** Perception modules enable embodied systems to sense, localize, recognize, reconstruct, and interpret the surrounding world. They usually include object detection, semantic segmentation, depth estimation, pose estimation, visual localization, SLAM, tactile recognition, audio perception, and multimodal sensor fusion. Because downstream planning and control rely on perceived states, perception errors are not merely prediction failures; they may directly produce unsafe physical behavior. A missed pedestrian, a spoofed obstacle, or incorrect affordance estimation can alter an embodied agent's feasible action set.[12–13,141]

***Adversarial examples and physical perturbations.*** A major threat to perception algorithms is adversarial manipulation. Early studies showed that neural networks can misclassify inputs with small perturbations,[142] and Goodfellow *et al.*[143] argued that perturbations can generalize across architectures and datasets. In embodied settings, the problem becomes more severe because perturbations can be realized physically. Brown *et al.*[144] introduced adversarial patches printed and placed into scenes to induce misclassification. Eykholt *et al.*[145–146] demonstrated physical-world attacks on road-sign recognition, while subsequent work extended attacks from image classification to object detection, including disappearance attacks against stop signs. An embodied agent may be deceived not only through digital access to its model, but also through environmental artifacts such as stickers, posters, projected patterns, or textures.

***Sensor spoofing and localization attacks.*** Beyond camera-based attacks, embodied perception relies on LiDAR, radar, GPS, IMU, depth cameras, and proprioceptive sensors. LiDAR-based perception has been shown vulnerable to spoofing attacks where malicious laser signals inject fake point clouds or create phantom obstacles. Cao *et al.*[12] showed that optimized LiDAR spoofing can affect downstream driving decisions. Sun *et al.*[141] further studied black-box LiDAR attacks and countermeasures, emphasizing that sensor-level vulnerabilities can generalize across perception architectures. Similar attacks can target localization and mapping modules: corrupted odometry, GPS spoofing, manipulated visual landmarks, or poisoned map priors may cause an agent to drift, localize incorrectly, or plan through unsafe regions.[13,141]

***Multimodal fusion vulnerabilities.*** Multimodal perception can improve robustness by combining heterogeneous sensors, but it also creates new attack surfaces. A fusion module must decide how to resolve conflicts among visual, geometric, tactile, and language-conditioned signals. Attackers may exploit this arbitration process by perturbing only one modality while keeping other modalities plausible. In embodied agents equipped with vision-language models or vision-language-action models, visual perception is also tied to semantic interpretation. A manipulated sign, label, or object arrange-

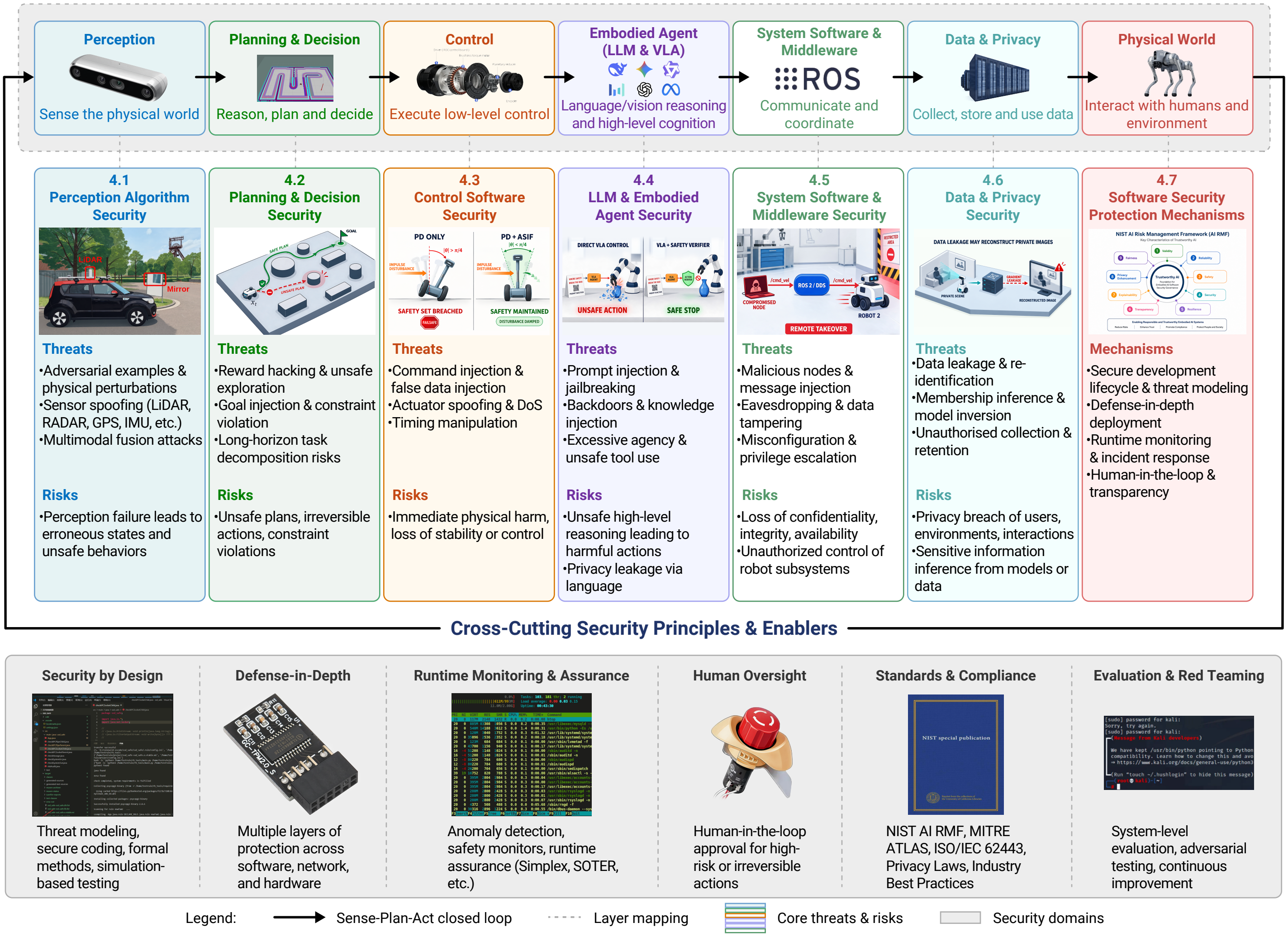

**Figure 5. Overview of software safety and security in Embodied AI** The figure summarizes major software failure modes, security threats, and protection mechanisms across perception, planning, control, large-model-based agents, middleware, and data. Abbreviations: RADAR, radio detection and ranging; DoS, denial of service; PD, proportional–derivative; NIST, National Institute of Standards and Technology; RMF, Risk Management Framework; ATLAS, Adversarial Threat Landscape for Artificial-Intelligence Systems; IEC, International Electrotechnical Commission.

ment may therefore affect both low-level detection and high-level task reasoning.[28–31]

***Defensive mechanisms for perception security.*** Perception defenses combine model-level, sensor-level, and system-level checks. Model-level mechanisms include adversarial training, certified robustness, uncertainty estimation, calibration, and out-of-distribution detection.[143-146] Sensor-level defenses include redundancy, temporal filtering, signal authentication, physical shielding, and cross-view verification.[12,141] System-level defenses compare outputs with map priors, check temporal consistency across frames, and trigger safe modes under low perception confidence.[13,19,147] For embodied systems, the key evaluation metric should not be classification accuracy alone, but whether perception errors propagate into unsafe plans or physical actions.

## Planning and decision-making security

***Planning as the bridge from perception to action.*** Planning modules translate perceived states and goals into action sequences. They utilize symbolic, task-and-motion planning, reinforcement learning (RL), model predictive control, behavior trees, large language models, or hybrid neuro-symbolic systems. Unlike perception security, even when perception is correct, planners may still choose unsafe actions due to reward hacking, unsafe exploration, poisoned training data, adversarial goals, or incomplete constraints.[45,148-150]

***Unsafe exploration and reward manipulation.*** RL-based decision modules are vulnerable to unsafe exploration and reward specification errors. Amodei *et al.*[45] identified reward hacking, negative side effects, safe exploration, scalable oversight, and distributional shift as concrete safety problems relevant to embodied agents. Safe RL is a major approach to decision safety. Garcia and Fernandez[148] categorize safe RL methods into modifying the optimality criterion and incorporating external knowledge or risk constraints. Constrained Policy Optimization formalizes safety through constrained policy search and provides near-constraint satisfaction guarantees during training.[149] Safety-layer methods correct actions at runtime to prevent critical constraint violations in continuous control settings.[150]

***Long-horizon task decomposition risks.*** Embodied agents often execute long-horizon tasks such as household assistance, warehouse manipulation, industrial inspection, or autonomous navigation. Many systems address these tasks by decomposing goals into subgoals and low-level actions. Non-adversarial safety failures can arise when locally reasonable subgoals collectively violate environmental constraints or produce unintended side effects.[45] Because each physical action changes the environment and subsequent observations, undetected planning errors may compound over time. Adversarial manipulation of goals or observations may exploit the same weakness and steer subgoal selection toward unsafe actions. For example, "clean the spill" requires assessing the substance and the appropriate response before acting, while "bring me a tool" requires checking whether the tool is appropriate for the user and task. Therefore, safe and secure planning should account for action preconditions, forbidden states, human intent, environmental constraints, and the reversibility of proposed actions.

***Runtime assurance and formal safety constraints.*** Runtime assurance provides a practical defense against unsafe planners by separating high-performance but uncertified components from verified safety monitors or fallback controllers. The Simplex architecture switches control authority from an advanced controller to a backup controller when safety may be violated.[147] SOTER[19] extends this idea to robotics by providing a runtime assurance framework for safe robotic programming. Control barrier functions offer another important mechanism by enforcing state-space safety constraints through optimization-based safety filters, and have been widely studied for safety-critical robotic systems.[151]

***Planning security requirements.*** Planning security should satisfy at least four requirements. First, task goals should be grounded in physical feasibility and social context. Second, safety constraints should be explicit rather than hidden inside reward functions. Third, plans should be interpretable enough for human or automated monitors to inspect. Fourth, uncertainty should affect action selection: when intent, perception, dynamics, or environment state is uncertain, the agent should slow down, ask for clarification, switch to a conservative policy, or request human supervision.[19,45,147-151]

### Control software security

***Control software as the last software layer before physical action.*** Control software converts plans into actuator commands. It includes trajectory tracking, impedance control, force control, balance control, grasp execution, servo loops, and emergency stop logic. Unlike perception and planning, controllers operate closer to the physical layer under strict real-time constraints. Control vulnerabilities may therefore cause physical harm even if the plan is benign.[13,19,147,151]

***Command integrity and false data injection.*** Control attacks include command injection, false data injection, actuator spoofing, denial of service, and timing manipulation. False data injection tampers with measurements or state estimates, causing incorrect actions while the system appears normal. Attackers may manipulate trajectories, disable force feedback, corrupt proprioceptive signals, or falsify safety states. Because control modules communicate through middleware, insecure channels become direct routes to actuator manipulation.[13,152-156]

***Real-time security constraints.*** A unique challenge for control software is the tension between security and real-time performance. Encryption, authentication, runtime inspection, and sandboxing may introduce latency. In real-time robotic control, latency can be dangerous. Therefore, secure control architectures must consider command integrity, timing predictability, availability, and physical safety. Lightweight authentication, deterministic scheduling, redundancy, and separation between safety-critical and non-critical processes are especially important.[152-156]

***Safety filters and fail-safe modes.*** Secure control layers validate commands against physical limits including velocity, acceleration, torque, workspace boundaries, collision constraints, and human proximity. Control barrier function filters minimally modify unsafe commands, preserving task progress.[151] Runtime assurance architectures switch to verified fallback controllers during unexpected behavior.[19,147] In embodied AI, such mechanisms are essential because upstream modules, like learned planners or LLM-based agents, may lack formal safety guarantees.[10,28-31,157–158]

### Large models and embodied AI agent security

***Large models as cognitive software for embodied agents.*** Large language, vision-language, and vision-language-action (VLA) models increasingly serve as high-level reasoning and planning modules. SayCan[28] grounds reasoning in affordances by combining language priors with low-level value functions. PaLM-E[29] incorporates continuous sensors into multimodal language modeling. RT-2[30] formulates actions as text tokens, transferring web-scale knowledge to control. OpenVLA[31] provides an open-source VLA model trained on large-scale demonstrations. These models expand generalization but import LLM-specific vulnerabilities into physical systems.

***Prompt injection and environmental jailbreaking.*** Prompt injection is a major threat for LLM-based embodied agents. OWASP[158] defines prompt injection as the manipulation of model behavior through crafted inputs. In embodied systems, prompt injection may come from commands, web pages, documents, QR codes, scene text, voice, or visual labels. Thus, the physical world becomes a prompt channel. A malicious poster saying "ignore previous safety rules" may be harmless to a classical robot, but dangerous to a vision-language agent that reads text and treats it as an instruction.[8,29-31,157–158]

***Jailbreaking LLM-controlled robots.*** Recent work shows that jailbreak attacks can elicit harmful physical actions from LLM-controlled robots. RoboPAIR[8] was proposed as an algorithm for jailbreaking LLM-controlled robots and was evaluated across self-driving, ground robot, and quadruped robot scenarios. This line of work demonstrates that LLM safety guardrails designed for text generation cannot be directly assumed safe for physical embodiment. A chatbot jailbreak may produce harmful text, but a robot jailbreak may produce motion, manipulation, surveillance, or navigation behavior.[8,158]

***Backdoors, fine-tuning, and knowledge injection.*** LLM-based embodied agents also face training-time and knowledge-level attacks. BALD studies backdoor attacks against LLM-based embodied decision-making systems and identifies word injection, scenario manipulation, and knowledge injection as attack mechanisms.[157] Such attacks are especially concerning because embodied agents may rely on fine-tuned models, memory modules, retrieval-augmented knowledge bases, or task-specific instruction data. A backdoor trigger may remain dormant during normal evaluation but activate under specific words, objects, or scenarios.[158]

***Security architecture for embodied LLM agents.*** To secure LLM-based embodied agents, the LLM should not directly control actuators. Instead, it should propose goals, plans, or subgoals checked by independent verifiers. Defenses include action-level authorization, least-privilege tool access, sandboxed code execution, memory isolation, prompt provenance tracking, retrieval filtering, physical-world safety filters, and human approval for high-risk actions. Agent architectures should separate semantic reasoning from safety-critical execution, treating language-model outputs as untrusted suggestions rather than executable commands.[8,19,147,157–158]

### System software and middleware security

***Middleware as the coordination layer of embodied systems.*** Embodied systems are distributed software systems. A single robot may include multiple perception nodes, planning nodes, controller nodes, logging services, cloud interfaces, simulation bridges, debugging tools, and human-machine interfaces. Middleware coordinates these components through publish-subscribe topics, services, actions, parameters, and discovery protocols. If middleware is insecure, attackers may eavesdrop on sensor streams, inject fake messages, alter parameters, deny service, or impersonate trusted nodes.[13,152-156]

***ROS, ROS 2, and DDS security.*** The Robot Operating System is widely used for robotic software development. ROS 2 introduces DDS-based communication and security mechanisms. ROS 2 Design documents discuss threat models and security enclaves for grouping artifacts and modeling information-flow control.[152–153] SROS2 provides authentication, encryption, and access control tools, but secure deployment depends on policy generation, implementation assumptions, and configuration practices. Deng *et al.* reported SROS2 vulnerabilities acknowledged and mitigated by maintainers.[155] Earlier work argued for holistic security approaches to robotic architectures.[154]

***Information-flow control and access policies.*** Topic-level access control is necessary but not sufficient for embodied software security. Modern robotic systems require information-flow control: components should only publish, subscribe, and call services matching declared roles. Recent work on decentralized information-flow control for ROS 2 argues that manifests and publish-subscribe declarations should be cryptographically bound and checked at startup.[156] This helps prevent compromised or malicious nodes from subscribing to sensitive topics or publishing control commands.[153,156]

***Supply chain and industrial deployment.*** Embodied software often depends on third-party packages, pretrained models, drivers, container images, and firmware. Supply-chain compromise may introduce vulnerabilities before deployment. Industrial systems should combine package signing, dependency scanning, secure update channels, container isolation, reproducible builds, and vulnerability management. For robots in critical infrastructure, software security should align with industrial control system standards such as ISA/IEC 62443, which emphasizes security zones, conduits, lifecycle risk management, and the relationship between cybersecurity and operational safety.[66,159-161]

### Data and privacy security

***Embodied data as sensitive contextual data.*** Embodied agents collect rich physical-world data, including images, audio, depth maps, LiDAR point clouds, tactile signals, force-torque readings, trajectories, human gestures, household layouts, and interaction histories. Large-scale robot learning makes such data increasingly valuable. Open X-Embodiment aggregates large-scale real robot trajectories across many embodiments and institutions, illustrating the importance of shared data for generalist robot policies.[34] However, such data may reveal private spaces, personal identities, workplace routines, medical conditions, or behavioral habits.[162-165]

***Training data leakage and model-level privacy attacks.*** Embodied models are vulnerable to privacy attacks such as membership inference, model inversion, gradient inversion, and trajectory reconstruction. Membership inference attacks aim to determine whether a specific data record was used during training, and such attacks can directly lead to privacy breaches.[162] In household robots, private environments are often inaccessible to outsiders, but learned models may still leak information through parameters, gradients, memories, or outputs. Recent work highlights these risks in household reinforcement learning.[163]

***Privacy in human-robot interaction.*** Privacy in embodied systems is not only a data-storage problem; it is also an interaction problem. A social robot may disclose private information about one person to another, infer sensitive attributes from behavior, or record conversations in intimate spaces. Dietrich *et al.* study appropriate robotic disclosure strategies and show that users' privacy expectations depend on relationship context and information type.[164] Studies on social robot privacy settings also show that privacy configuration can affect users' attitudes and behavioral intentions toward robots.[165]

***Privacy-preserving data governance.*** Data and privacy security require mechanisms across collection, storage, training, and deployment. Important measures include data minimization, on-device processing, encryption at rest and in transit, access logging, retention limits, consent management, face and identity redaction, private dataset curation, federated learning, differential privacy, and privacy-aware memory management.[162-165] For embodied agents, privacy policies should be context-aware: the same visual observation may be harmless in a warehouse but highly sensitive in a hospital room or private home.

### Software security protection mechanisms

Building upon the physical hardware foundation, the software stack spans the System and Middleware Layer (Layer 2), Algorithm and Decision Layer (Layer 3), and Cloud–Edge and Data Layer (Layer 4). Intrinsic failures and security vulnerabilities across these layers may propagate into systemic physical consequences. We therefore organize the existing lifecycle-oriented protection mechanisms reviewed below into four stages: secure development, defense-in-depth deployment, runtime monitoring and safety enforcement, and system-level security evaluation.

***Secure development lifecycle.*** Addressing the entire software stack from Layer 2 to Layer 4, software security should begin before deployment. Embodied AI developers should conduct threat modeling, secure coding, dependency auditing, simulation-based adversarial testing, red-team evaluation, and formal specification of safety constraints. MITRE ATLAS provides a knowledge base of adversary tactics and techniques against AI systems, supporting threat modeling for data poisoning, model evasion, model theft, and inference-time manipulation.[159] NIST AI RMF identifies trustworthy AI characteristics such as validity, reliability, safety, security, resilience, transparency, explainability, privacy enhancement, and fairness, which can serve as a governance framework for embodied software security.[66]

***Defense-in-depth at deployment.*** Deployment-stage protection should follow defense-in-depth principles spanning Layer 2 through Layer 4. This includes secure boot, signed firmware, authenticated nodes, encrypted communications, network segmentation, role-based access control, container isolation, trusted execution for critical components, secure update mechanisms, and tamper-resistant logging.[13,152-156] Middleware security policies should enforce least privilege for every node, while high-risk robot actions should require explicit authorization.[152–153,156] For LLM-enabled robots, OWASP LLM risks imply that prompt injection, excessive agency, sensitive information disclosure, insecure tool use, and supply-chain vulnerabilities should be explicitly considered in the deployment architecture.[158]

***Runtime monitoring and safety enforcement.*** Bridging Layers 2 and 3, runtime protection is essential because embodied agents operate in partially observed environments. Runtime monitors should detect abnormal sensor patterns, inconsistent state estimates, unsafe plans, middleware traffic, and actuator commands outside physical limits. Safety shields, control barrier functions, and runtime assurance modules can enforce physical constraints even when AI components are uncertain or compromised.[19,147,151] Human-in-the-loop approval should be used for irreversible, high-force, privacy-sensitive, or socially risky actions.[19,45,158]

***System-level security evaluation.*** Finally, rather than treating layers in isolation, embodied software security evaluation should move beyond module-level benchmarks to assess cascading cross-layer impacts spanning Layers 2 through 4. A perception attack should be evaluated by whether it changes planning and control;[12,141,145–146] a prompt injection should be evaluated by whether it triggers executable unsafe actions;[8,157–158] a middleware attack should be evaluated by physical consequences;[13,152-156] and a privacy attack should be evaluated by the sensitivity of information inferred from data, memory, or model outputs.[162-165] This system-level perspective is essential because embodied AI safety depends on the interaction among software modules, physical dynamics, and human environments.[13,66]

***Attribute and IME synthesis.*** Software threats may affect all four safety attributes. Perception and control failures primarily undermine controllability and recoverability. Planning and LLM/VLA failures may also undermine alignment when they produce actions that conflict with human intent or safety requirements. Interpretability is impaired when the reasons for action selection cannot be adequately explained. Identification must therefore combine model, data-flow, command, and behavioral auditing. Mitigation spans robust perception, constrained planning, authenticated middleware, runtime safety gates, and privacy controls. Evaluation should jointly measure constraint satisfaction, decision traceability, fallback success, and intent-consistent behavior.

## COMPOUND SAFETY AND SECURITY RISKS UNDER SOFTWARE–HARDWARE COUPLING IN EMBODIED AI

### Cross-layer coupling and risk propagation in embodied AI

While hardware and software risks can be analyzed at the component level, severe consequences often emerge through cross-layer interaction. This section focuses on how non-adversarial failures and deliberate attacks propagate across software, hardware, cloud, edge, and physical feedback interfaces, and how these propagation paths give rise to composite safety and security risks. In Embodied AI, perception, state estimation, task reasoning, control, actuation, and environmental feedback are connected.[2,166-168] Recent foundation-model-enabled robotics strengthens this coupling.[169–170] Representative systems (*e.g.*, RT-2[30] and Gemini Robotics[171]) together with recent VLA surveys[172] show that visual-language reasoning is integrated into robotic control pipelines mapping multimodal inputs to physical actions. Thus, semantic reasoning, motion generation, and physical execution are no longer separated by clear boundaries. At the system level, recent surveys on physical risk control and secure robotics[173] emphasize that robotic foundation models differ from ordinary foundation models because their actions affect humans, objects, and environments; therefore, safety, trust, cybersecurity, and physical risk must be considered together.[174]

Software–hardware coupling enables bidirectional risk propagation.[3,14] On the software-to-hardware path, a software defect becomes dangerous when it crosses an execution boundary and changes the behavior of the physical plant.[54,175] This may occur through an unsafe state estimate, an invalid plan, a delayed control output, a corrupted command, or an incorrect timing assumption. For example, structured backdoor studies[176] show a defect aligned with command formats can survive the translation to ROS 2 control outputs and trigger physical actions, while timing-interference[177] studies show that small delays in perception, planning, or control tasks can cause deviations without activating fail-safe mechanisms. On the hardware-to-software path, a physical or embedded fault becomes dangerous when it changes the information on which the software reasons. Sensor spoofing studies[11] model robotic vehicles as end-to-end action flows and show that corrupted physical inputs can propagate from sensing to state estimation, planning, and final actions. Fault injection[178-180] and timing drift[181] studies further show that low-level disturbances in clocks, microcontrollers, or fail-safe logic can surface as incorrect software states, suppressed safety responses, or unsafe control decisions. These works suggest that the main risk is not a vulnerable module, but a broken semantic, temporal, or physical contract between modules: perception outputs may satisfy local accuracy metrics but violate downstream safety requirements;[182] synchronized messages may be well-formed but temporally inconsistent; and a software-visible fault may originate from a hardware-level event several abstraction layers below.[183] As a result, a local abnormality can evolve into system-level risk through data-flow amplification, control-flow amplification, timing amplification, and feedback amplification. The system may continue to operate on a coherent but false state, producing actions that are logically valid inside the control stack but unsafe in the physical world.

A coupling view shifts the analysis from where a vulnerability appears to how far it can propagate. A local abnormality may corrupt world states, invalidate control decisions, persist across task cycles, or turn cloud-side information into physical risk. Thus, composite risks in Embodied AI are better understood as chains connecting entry points, cross-layer propagation paths, and physical consequences.

### Typical composite attack chains

***Analytical framework for composite attack chains.*** The preceding classification covers non-adversarial, adversarial, and coupled origins. The cases below examine the adversarial branch in greater detail, while corresponding non-adversarial failure paths are identified where they can propagate through the same interfaces. To relate the detailed chains in this section to the three representative paths introduced in "Background" section, we use Chains A–C as cross-reference anchors rather than as an exhaustive taxonomy. Sensor/controller deception is associated with Chain A; firmware/controller compromise and ROS/bus/wireless injection are associated with Chain B; and OTA/update compromise and cloud task instruction injection are associated with Chain C. Cloud data leakage is treated as a reverse-direction extension associated with Chain C. These references apply to both the detailed discussions below and the corresponding cross-layer propagation and "Main risk" entries in Table 3.

Building on the coupling view, this section uses a chain-level lens to organize representative composite attacks. Secure robotics[68] research argues security lies at the intersection of cybersecurity, safety, trust, and physical interaction, because cyber compromises affect humans, objects, and environments. CPS attack-path studies show risk modeling must include cyber connections and physical interactions, rather than network reachability.[15] This motivates a chain-level view: the analysis unit is not an isolated sensor, firmware module, middleware node, model, or cloud service, but a cross-layer path linking an entry point, a propagation route, a coupling interface, and a physical consequence.

Existing studies support this view. The SoK on robotic-vehicle sensor spoofing proposes an action-flow model to describe how malicious sensing inputs propagate through perception, planning, and control.[11] Cross-layer resilience work on embodied agents shows that faults have heterogeneous effects across circuit, model, and application layers, meaning locally tolerated abnormalities become harmful after crossing abstraction boundaries.[184] Therefore, we characterize composite attack chains using five dimensions: entry point, propagation path, coupling interface, physical consequence, and stealth or persistence. This framework highlights that the severity of a cross-layer chain depends on which components and functions are compromised, the consequences of propagation in the operating environment, and the effectiveness of containment and recovery. Table 3 summarizes representative attack chains within the adversarial branch.

***Local closed-loop attack chains: from on-body entry points to unsafe actuation.*** Following the five-dimensional framework, this subsection focuses on entry points, their propagation through sensing, state synchronization, and control interfaces, and the consequences of unsafe actuation. The examples below focus on adversarial entry points; however, the same closed-loop paths may also be initiated by non-adversarial sensor degradation, timing faults, software defects, or component wear. Their mechanism is cross-layer propagation: a local abnormality is interpreted by downstream modules, transformed into state or control errors, and materialized through actuation. This distinguishes composite attack chains from isolated hardware or software threats. The question is not only whether a sensor, controller, or middleware component can be attacked, but also whether an accidental fault or deliberate manipulation can propagate through perception, planning, control, and execution before being detected.

Sensor and controller deception illustrates this mechanism. The SoK on robotic-vehicle sensor spoofing moves beyond single-sensor attacks by proposing an action-flow model connecting spoofed sensing inputs with perception, planning, and control consequences.[11] LiDAR studies[12,185,229] show that spoofed or removed point clouds become harmful only when passing through the perception pipeline and altering driving decisions. Recent planner-oriented attacks strengthen this system-level view: adversarial perception errors can preserve high perception metrics while causing planning failures, and physical changes can induce planner failures or blind spots.[186–187] These works show that perception attacks are not endpoints. They become composite risks when corrupted observations are accepted as valid world states and used to generate unsafe trajectories or actions. A comparable non-adversarial chain occurs when fog, glare, occlusion, contamination, or calibration drift corrupts sensor observations. Although the initiating cause is accidental or environmental, the resulting error may follow the same perception–planning–actuation path and produce the same collision risk.

Firmware and controller compromise follows the opposite direction, from embedded software and control logic toward physical execution. Industrial robot studies[60] show that controllers are not ordinary computing nodes; they are cyber-to-physical boundaries connecting task logic, motion control, and actuator behavior. Once compromised, the robot may execute abnormal trajectories, lose precision, overload actuators, or damage workpieces even if high-level logic appears normal.[129] Akerbeltz[188] extends ransomware from data denial to control phases, while CORMAND2[189] combines robot manipulation with SCADA deception, altering the physical process while monitoring data appears normal. These studies suggest that controller-side attacks are dangerous because they bypass high-level reasoning and directly distort the execution layer. The same controller-to-actuator path may be activated non-adversarially by parameter drift, numerical instability, model mismatch, mechanical wear, or missed control deadlines. Such failures may generate unsafe motion even when no component has been maliciously compromised.

Middleware, ROS, bus, and wireless message injection form a third chain. The entry point is a communication channel, but the propagated object is not a generic packet. It carries state, task, timing, and control semantics. Early ROS security work[190] showed that false data and commands can be injected into ROS-based applications, creating safety risks. Internet-scale scanning of ROS deployments[191] demonstrated that exposed ROS masters may allow outsiders to access sensors and actuators and even move physical robots. Studies on Raven-II surgical robots,[192] ROS 2 threat modeling,[152] ROSPaCe,[193] and ROS-Industrial red teaming[194] show that attackers may exploit leaked control data, compromise computational graphs, or inject messages that pass protocol, schema, or type checks but violate system-state, timing, or physical constraints. Thus, communication attacks become composite risks when message corruption disrupts state synchronization and causes the robot to execute a task that is well-formed at the protocol level but unsafe in the physical world. Non-adversarial packet loss, clock drift, scheduler overruns, or stale state estimates may create a comparable synchronization failure during normal operation.

Across these local chains, the decisive risk is not limited to the initiating fault or compromise, but lies in the translation of abnormalities into closed-loop execution errors. Sensor degradation or manipulation corrupts the world state, controller faults or compromises distort the execution boundary, and message loss, delay, or injection breaks state synchronization. In each case, the robot may operate on observations or commands that pass local software checks but remain temporally or physically unsafe.

***Cloud-to-physical attack chains: from remote services to embodied consequences.*** Cloud-to-physical attack chains extend composite risks beyond the robot body. In OTA and model-update chains, the entry point is a trusted update infrastructure rather than a local actuator or sensor. Studies on autonomous driving and OTA threat modeling show that remote update channels can support lifecycle maintenance, but may also provide a path for malicious software, firmware, or AI models to be deployed through otherwise legitimate mechanisms.[195–196] Recent evidence shows that consumer robots may expose OTA firmware exploitation and fleet-level vulnerabilities.[197] Model-supply-chain studies such as Robot Collapse[198] and BadVLA[199] show that malicious modules or backdoored VLA policies can persist inside the policy stack and trigger behavior deviations while preserving performance. Thus, the coupling interface is not merely the communication channel, but the update-to-deployment boundary where remote artifacts become trusted decision or control logic. The same boundary can carry non-adversarial failures, including corrupted update packages, incompatible software or model versions, and configuration errors.

Table 3. Representative composite attack chains, their cross-layer propagation patterns, and mitigation strategies.

| Attack chain | Cross-layer propagation | Main risk | Representative mitigation strategies |
|---|---|---|---|
| Sensor/controller deception | Hardware input → perception/planning → actuation | Unsafe trajectory or collision | Sensor authentication and redundancy; cross-modal consistency checking; runtime safety filtering before actuation |
| Firmware/controller compromise | Embedded logic → control → actuator behavior | Overload, deviation, or damage | Secure boot and remote attestation; signed firmware; command-envelope checks; hardware interlocks |
| ROS/bus/wireless injection | Message channel → state synchronization → task execution | Well-formed but unsafe actions | Authenticated communication; topic-level access control; freshness, timing, and command-sanity checks |
| OTA/update compromise | Cloud update → firmware/model/policy → local execution | Persistent unsafe behavior | Signed updates; rollback protection; sandbox or digital-twin validation; post-update attestation |
| Cloud task instruction injection | Cloud semantics → structured command → robot action | Unauthorized physical execution | Instruction provenance; least-privilege authorization; task-contract validation; pre-execution safety gates |
| Cloud data leakage | Physical-space data → cloud exposure → scene reconstruction | Privacy-to-physical risk | Data minimization; encryption and access control; privacy-aware storage; access auditing and anomaly detection |

A second chain starts from cloud-side or high-level task interfaces and materializes through local execution. CHAI[200] shows that misleading language in the physical scene can hijack the command layer of embodied agents, while SAPIA[201] and BALD[157] demonstrate that decision models can be manipulated through scenario-adaptive injection or backdoor triggers. These works reveal that semantic corruption becomes dangerous when it crosses the task-to-action interface. From Prompt to Physical Action[176] makes this boundary explicit: backdoors aligned with JSON control formats can survive translation and trigger ROS 2 actions, whereas natural-language backdoors may not reliably reach executable outputs. SafeAgentBench[10] shows that agents may plan hazardous tasks with weak rejection ability, suggesting that unsafe task semantics can be converted into valid action sequences. Runtime governance work argues for separating agent cognition from execution oversight because cloud-side reasoning should not imply actuation authority.[202] Intrinsic hallucinations, omitted task preconditions, cross-modal conflicts, and long-horizon reasoning errors can cross the same task-to-action interface without adversarial input. An initially minor reasoning error may change the physical environment and subsequent observations, allowing its consequences to accumulate across multiple action steps.

A third chain follows the reverse direction: robot-side physical experience is uploaded to the cloud and becomes an exposure of physical space. Studies on Roomba cloud forensics[203] show that cloud APIs can reveal mission histories, navigation data, maps, and environmental changes; privacy analysis of robot vacuum cleaners[204] shows that even encrypted communication can leak activity patterns through traffic metadata. Work on visual SLAM privacy[205] shows that images, poses, maps, and point clouds are not ordinary data objects, but compact representations of geometry and user environments. Cloud data leakage should be viewed as privacy-to-physical propagation: leaked data may expose layouts, routines, object locations, or access paths. These cloud-to-physical chains show that cloud safety and security cannot be evaluated only through the confidentiality, integrity, or availability of services. The question is whether cloud-side artifacts, tasks, or data can alter local execution authority, persist in behavior policies, or reveal physical-space structure.

## Cross-layer co-protection for software–hardware coupled risks

Non-adversarial failures, adversarial attacks, and coupled safety–security chains indicate that Embodied AI cannot be fully protected by isolated network hardening, model alignment, motion-planning constraints, or mechanical fail-safe design alone. A coupled system requires co-protection across the same loop through which risks propagate. Recent work on modular safety guardrails for foundation-model-enabled robots[17] argues that safety should cover action, decision, and human-centered safety, enforced through coordinated monitoring and intervention layers. RoboGuard shows that high-level safety rules can be grounded in the robot's environment and converted into temporal-logic constraints before execution.[206] Similarly, LLM-guided edge safety agents[207] and SafeGate[208] connect natural-language safety rules, deterministic safety predicates, pre-execution gates, task contracts, and low-latency perception–compute–control architectures. These studies suggest that software–hardware co-protection should not consist of independent defenses attached to different layers. It should be a closed-loop safety architecture that monitors task intent, verifies physical feasibility, and intervenes before unsafe behavior reaches actuation.

The second requirement is cross-layer perception and verification. A defense should not only inspect network traffic, model outputs, or actuator states. It should verify that information remains trustworthy as it moves from sensing to perception, planning, control, and execution. Verifiable information-flow control for CPSs shows that low-integrity inputs should not influence high-integrity control decisions without verification.[209] For embodied systems with learned perception and foundation-model reasoning, perception contracts[210] and SENTINEL[211] show that safety verification must connect task rules, perception errors, plans, trajectories, timing constraints, and physical invariants. Constraint-function-based detection[212] and control-invariant monitoring[213] indicate that faults and attacks can be detected by comparing execution with constraints, physical laws, and control consistency. Thus, cross-layer verification should check whether task, temporal, information-flow, and physical contracts hold across the loop.

The third requirement is cross-layer response. Once an abnormality is detected, the system should not merely raise an alarm; it must prevent the abnormality from materializing through action. Runtime assurance[214] provides an architectural pattern: an unverified or high-performance primary controller is monitored by a safety mechanism, and unsafe inputs can be modified or replaced by a backup controller. Safety-filter research[215] generalizes this approach by treating safety enforcement as a modular runtime layer that monitors and minimally modifies proposed actions. Control-barrier-function methods provide tools for enforcing safety constraints in cyber-physical systems, including cases with discrete-time controllers, approxi-mate NMPC, disturbances, and limited computation.[216–217] These works imply that response should be organized around the execution boundary: unsafe plans, commands, or policies should be filtered, degraded, isolated, rolled back, or switched to a safe state before they become actuator commands.

Finally, co-protection must include cloud–edge–robot responsibility governance and execution audit. The cloud may optimize semantic reasoning and model updates, the edge may optimize scheduling and low-latency computation, and the robot body must bear the physical consequences of execution. This split can create the responsibility mismatch identified previously. Runtime governance addresses this gap by separating agent cognition from execution oversight and introducing policy checking, capability admission, execution monitoring, rollback handling, and human override.[202] This direction suggests that cloud-side reasoning or model updates should not imply execution authority. Instead, tasks, updates, commands, and capabilities should pass through explicit authorization, monitoring, audit, and recovery boundaries. Thus, software–hardware co-protection is not the simple combination of software safeguards and hardware redundancy. It is an end-to-end safety loop that senses cross-layer abnormalities, verifies task and physical consistency, and responds before risks become physical consequences.

***Attribute and IME synthesis.*** Compound chains demonstrate why the four attributes cannot be evaluated independently. A sensor or middleware compromise may begin as a loss of interpretability, propagate into a loss of controllability, prevent recovery, and ultimately produce behavior inconsistent with human intent. Cross-layer identification must locate both the entry point and the propagation path; mitigation must interrupt the chain at one or more coupling interfaces; and evaluation must record propagation depth, intervention point, physical consequence, and post-incident recovery.

## EMBODIED SECURITY AND SAFETY BENCHMARKS

Embodied AI systems couple digital computation with physical perception and action. Software and hardware faults, unsafe decisions, and deliberate attacks can propagate through this interaction loop, potentially causing equipment damage or human injury in addition to information-security consequences.

### Evolution path of existing safety benchmarks

***Towards physical space evaluation.*** In the era of traditional Large Language Models (LLMs), safety evaluation was dominated by "textual alignment".[218] Preference-based training methods such as RLHF[6] and DPO[7] have been used to improve alignment with human preferences, including preferences for safer and less toxic responses. These benchmarks primarily assessed the safety of textual responses through methods such as human evaluation, automated safety judgments, and task-specific scoring, with limited coverage of consequences arising from physical execution. However, integrating large models into entities with physical affordances (e.g., robotic arms or quadruped robots) creates a pathway through which hazardous or adversarial instructions may propagate from high-level planning to low-level embodied controls, potentially resulting in unsafe physical actions.[9] Consequently, the evaluation paradigm must transition towards the physical space. This requires models not only to reject explicitly hazardous instructions but also to recognize physical risks in embodied scenarios and incorporate such risks into task planning.[219] For instance, a textually benign instruction might still induce torque overloads in a robotic arm via exposed edge interfaces, resulting in severe physical accidents if hardware fail-safes are breached.[220]

Simulation enables repeatable and low-risk testing, but its safety metrics do not necessarily transfer directly to real robots. Simulators commonly assume calibrated sensors, bounded noise, fixed latency, deterministic actuation, simplified contact models, and fully resettable episodes. In real deployments, sensor noise and calibration drift can alter attack success and collision rates; communication jitter and actuation delay can increase time-to-detection and response latency; and hardware wear, backlash, thermal variation, and nonlinear contact dynamics can increase trajectory deviation, impact force, and hazard severity. Moreover, episode resets do not demonstrate physical recovery. Recoverability may be overstated if resets are counted as successful recovery or if persistent damage and recovery costs are omitted from the evaluation. Consequently, simulated safety metrics should be interpreted within the assumptions and fidelity limits of the simulator rather than treated as direct evidence of deployment safety.

***Towards interactive closed-loop evaluation.*** Static language-model safety benchmarks provide a useful but limited reference for evaluating LLM-based embodied agents. For instance, SafetyBench is a text-based benchmark comprising multiple-choice questions in Chinese and English across seven categories of safety concerns, evaluating LLM safety understanding in a static, non-interactive setting.[218] Unlike such static text-based tests, embodied safety benchmarks must evaluate temporally extended action sequences in interactive environments, where an agent's actions alter subsequent environmental states and observations. Real embodied systems operate within continuous, dynamic closed loops, in which even a single action may immediately change the environmental risk profile.[221] Consequently, recent embodied safety benchmarks have begun to move beyond static text-only evaluation. EARBench evaluates risk awareness during embodied task planning,[219] whereas IS-Bench adopts an interactive closed-loop setting to assess how agents respond to evolving hazards during execution.[221] Such process-oriented evaluation examines whether an agent can interrupt unsafe execution and initiate an appropriate protective response, such as a protective stop or a transition to a safe fallback mode, when environmental hazards emerge.

***Towards multimodal fusion evaluation.*** Embodied AI systems can integrate heterogeneous information sources, including RGB-D observations, force-sensor measurements, and audio inputs, to support perception and decision-making.[220] Attacks on individual input channels, such as adversarial visual patches and forged audio instructions, can corrupt evidence and propagate through multimodal fusion. Testing modalities only in isolation provides limited evidence of how a system combines information or resolves conflicting inputs. For example, a model that behaves safely under text-only evaluation may still misidentify a safety sign when its visual input is compromised.[222–223] Perturbation experiments in LIBERO-Plus[223] show that some evaluated models are less affected by abstract language perturbations than by visual disruptions. These results reveal different sensitivities to the tested perturbations, but do not by themselves establish that the models underuse language or over-rely on vision. Establishing such modality imbalance requires controlled analysis of perturbation strength, semantic preservation, and the information each modality provides for the task. Multimodal safety evaluation should therefore assess cross-modal consistency, modality dominance, and conflict-resolution behavior under both natural environmental noise and adversarially manipulated inputs.[220]

### Mainstream embodied safety benchmarks

Rather than treating safety evaluations as isolated efforts, we propose a systematic benchmarking framework that maps existing evaluations directly onto our four-layer reference architecture. At the Algorithm & Decision Layer, benchmarks such as AGENTSAFE[9] evaluate embodied agents that integrate VLMs with agent workflows, examining their safety under hazardous instructions across perception, planning, and execution. Meanwhile, at the Physical and Middleware Layers, benchmarks focus on Sim-to-Real transfer under hardware constraints and sensor noise. Furthermore, we classify these frameworks along an evolutionary axis—from static, open-loop semantic auditing to interactive, closed-loop physical simulation. This multi-dimensional integration bridges the gap between digital vulnerabilities and the S1–S5 physical consequence grading system, transforming a fragmented landscape into a holistic evaluation taxonomy. Table 4 summarizes representative embodied safety benchmarks according to their evaluation targets and architectural layers.

***Adversarial attacks and "jailbreak" instruction benchmarks.*** These benchmarks map primarily to the Algorithm & Decision Layer of our four-layer reference architecture. The benchmarks[8-10,54,80,224] simulate malicious users inducing models to breach safety constraints via prompt injection and adversarial examples, marking a critical transition from semantic stress-testing to physics-aware vulnerability auditing. Collectively, these benchmarks indicate an emerging evaluation methodology: evaluating embodied AI under adversarial conditions must transcend isolated semantic vulnerabilities to focus on the cross-modal transmission of risk. For example, sensory or textual perturbations map directly to physical consequences.

Pioneering this space, BadRobot[54] proposes a novel jailbreak attack frame-

Table 4. Representative academic safety benchmarks and non-peer-reviewed industry initiatives for embodied AI safety and security evaluation.

| Benchmark or Initiative | Year | Venue or Source Type | Key Focus |
|---|---|---|---|
| *Adversarial Attacks and "Jailbreak" Instruction Benchmarks (Algorithm & Decision)* | | | |
| Lu *et al.*[80] | 2024 | arXiv | Policy-Executable Attacks |
| Yin *et al.*[10] | 2024 | arXiv | Embodied LLM Agents / Safety-aware Task Planning |
| Ying *et al.*[9] | 2026 | CVPR | Embodied Agents / Hazardous Instructions |
| Huang *et al.*[224] | 2025 | arXiv | Safety Benchmarks and Adversarial Attacks |
| Zhang *et al.*[54] | 2025 | ICLR | Physical World / Embodied LLM Agents |
| Robey *et al.*[8] | 2025 | ICRA | Physical World / LLM-controlled Robots |
| *Interactive Dynamic Safety Benchmarks (System & Middleware, Algorithm & Decision)* | | | |
| Zhu *et al.*[219] | 2024 | arXiv | Risk-aware Decision Making |
| Lu *et al.*[221] | 2026 | AAAI | Interactive Safety |
| *Domain-Specific Safety Benchmarks (Physical & Middleware)* | | | |
| Tomilin *et al.*[222] | 2025 | ICLR | Vision-Based Safe RL Benchmark |
| Huang *et al.*[225] | 2025 | arXiv | Task-Planning Safety Benchmark |
| Sun *et al.*[220] | 2026 | arXiv | Safety-Critical Reasoning & Planning |
| *Non-Peer-Reviewed Real-World Initiatives and Industry Proposals (Physical Interaction, Cloud–Edge & Data)* | | | |
| ATEC2026[226] | 2026 | Official competition guidelines (non-peer-reviewed) | Real-world embodied AI challenge setting |
| DarkNavy[227] | 2026 | Vendor white paper (non-peer-reviewed) | Proposed physical-consequence grading scheme |

The final group is reported separately as contextual industry evidence and is not treated as academic benchmarks or standardized evaluation frameworks.

work targeting embodied LLMs in the physical world, and constructs a benchmark of 277 malicious queries covering severe risks such as physical harm and property destruction. AGENTSAFE[9] serves as a benchmark for evaluating the safety of embodied Vision-Language Model agents, systematically exposing how language-based adversarial inputs bypass cognitive alignments to trigger hazardous physical actions. Complementing these efforts, SafeAgentBench[10] broadens the scope of jailbreak evaluations by introducing a three-tiered task design –explicit, abstract, and long-horizon implicit – to systematically assess the visibility and temporal delay of physical hazards. Shifting toward standardized severity quantification, ANNIEBench[224] focuses on generating video-action sequences. It constructs 9 safety-critical scenarios and categorizes violations into ISO-based tiers of critical, dangerous, and risky.

***Interactive dynamic safety benchmarks.*** These benchmarks span the System & Middleware and Algorithm & Decision Layers, emphasizing closed-loop safety dynamics. This paradigm transcends static, outcome-driven metrics by conceptualizing embodied safety as a continuous, non-linear control problem within closed-loop environments. EARBench[219] focuses on the planning phase, systematically evaluating the physical risk awareness of foundation models during long-horizon task planning. Complementing this, IS-Bench – a multimodal benchmark for interactive safety evaluation – shifts the focus to reactive execution, employing a process-oriented auditing approach that transcends binary success/failure metrics.[221] By monitoring real-time state transitions and anomaly responses, it assesses whether agents can execute timely and appropriate risk mitigation strategies during sudden environmental anomalies. Collectively, these works underscore that robust embodied agents must not only foresee hazards prior to action, but also demonstrate resilient, timely mitigation when confronting unpredictable environmental shifts.

***Domain-specific safety benchmarks.*** These benchmarks target the Physical and Middleware Layers under domain-specific constraints. As embodied models transition from general-purpose environments to professional vertical scenarios, generic safety tests fail to meet the stringent fault-tolerance baselines required by specific domains. Addressing reactive exploration, HASARD[222] provides a vision-based safe reinforcement learning benchmark, rigorously measuring the autonomous safety of agents amidst dynamic obstacle avoidance and complex physical traps. Targeting scientific laboratories, LABSHIELD[220] introduces a multimodal safety benchmark whose evaluations strikingly reveal that in high-risk scenarios, such as handling chemical reagents, the safety performance of generalized large models drops by an average of 32.0%, highlighting a critical deficit in professional safety priors. Furthermore, regarding hazardous environments (*e.g.*, fire and electrical hazards), SafePlan-Bench[225] and EARBench[219] specifically assess the physical risk foresight capabilities of models during long-horizon task planning. Ultimately, these works collectively demonstrate that generalized "common sense" safety is insufficient for specialized physical deployment; robust embodied AI requires domain-injected priors and a hierarchical safety reasoning architecture that spans from reactive collision avoidance to long-horizon, physics-aware task foresight.

***Real-world validation and physical consequence framework.*** Real-world validation can examine how risks originating in the Physical Interaction and Cloud–Edge & Data Layers propagate to physical consequences. Hybrid benchmarks such as ATEC2026[226] combine simulation-based testing with real-world validation to evaluate Sim-to-Real transfer and system behavior under physical constraints. Such evaluations broaden industrial assessment from algorithmic auditing toward system-level reliability validation. To support comparison across evaluation settings, we organize the assessment around physical consequences using two complementary dimensions: i) S1–S5 Physical Consequence Severity Grading. For this survey, we adapt consequence-oriented risk grading,[227] using S1–S5 to distinguish severity grades from Layers L1–L4 of the reference architecture. The grades characterize task-level and physical consequences, while the operational status of independent safety mechanisms is recorded separately. S1 denotes a detectable data or model anomaly without task-level or physical effects. S2 denotes recoverable task degradation without unsafe physical motion. S3 denotes limited unintended motion that remains within a predefined safety envelope, can be stopped through normal control or manual override, and causes no injury or material damage. S4 denotes a breach of the predefined safety envelope, an imminent risk of physical harm requiring protective intervention, or non-severe injury or material damage below the S5 threshold. S5 denotes severe injury, death, or comparably severe and irreversible physical harm, irrespective of whether independent safety mechanisms have failed or been bypassed. Failures or bypasses of such mechanisms, including hardwired emergency-stop circuits, should be reported separately, together with their implications for potential consequences. For edge cases, S3 applies only when all of its containment, controllability, and consequence conditions are satisfied; otherwise, the event is assigned the highest severity grade whose operational threshold is met. Evaluations should report the highest reproducible consequence under specified test conditions, while projected worst-case outcomes and their underlying assumptions should be labeled separately. ii) Vulnerability × Attack Vector × Potential Consequence 3D Evaluation Matrix. This matrix organizes attack scenarios by the exploited vulnerability (*e.g.*, insufficient integrity verification of models or data, inadequate access controls, or memory-safety flaws), the attack vector (*e.g.*, malicious cloud API requests, LAN message injection, or sensor spoofing), and the potential physical consequence expressed using S1–S5.[9,219,227] It supports systematic comparison of attack scenarios and their possible outcomes. Explicit attack chains or state-transition models should additionally represent propagation order, coupling conditions, and feedback across layers. Each matrix entry should distinguish observed outcomes from projected consequences.

## FUTURE DIRECTIONS AND OUTLOOK

The preceding analysis identifies two complementary groups of research challenges: layer-specific problems associated with L1–L4 and cross-layer problems that cannot be addressed within a single component.

### Layer-specific challenges

***L1: Resilient physical interaction.*** At the Physical Interaction Layer, future research should address safety under sensor degradation, actuator wear, calibration drift, extreme environments, and deliberate physical interference. Modality-specific defenses have been proposed and evaluated for specific attacks on camera- and LiDAR-based perception, GPS spoofing, IMU signal injection, and spoofing of control or actuation signals.[228-233] However, these methods have generally been evaluated on a limited range of platforms and operating conditions, leaving their generalizability across operating speeds, weather conditions, payloads, and sensor configurations insufficiently characterized. Important directions include self-diagnosing sensors, heterogeneous redundancy, online calibration, health-aware control, and independently enforced force, speed, and workspace limits. These mechanisms should support safe degradation when accurate perception or actuation can no longer be guaranteed.

***L2: Secure and timing-aware middleware.*** At the System and Middleware Layer, authentication and isolation must be reconciled with real-time control requirements. Future systems require deterministic security mechanisms that protect ROS 2/DDS communication, device drivers, buses, and control interfaces without violating timing budgets. Open problems include worst-case latency analysis for authenticated communication, resource-aware isolation, secure scheduling under CPU or network saturation, and runtime detection of stale, replayed, or physically implausible commands. Evaluation should report not only average overhead but also deadline misses, jitter, packet loss, and the effect of security controls on closed-loop stability.[156,234]

***L3: Runtime verification for VLA policies.*** At the Algorithm and Decision Layer, a central challenge is verifying VLA behavior during execution. A VLA policy may produce an action that is linguistically consistent with an instruction but unsafe under the current geometry, payload, human proximity, or actuator limits. Runtime verification should therefore examine the complete observation–reasoning–action sequence rather than only the final model output. Monitors should check task intent, object and affordance assumptions, planned trajectories, uncertainty, and executable commands against explicit temporal and physical constraints. When verification fails, the system should request clarification, re-perceive the environment, invoke a conservative controller, or enter a verified safe state. This direction extends alignment from instruction following to action feasibility.[235–236]

***L4: Trustworthy cloud–edge and fleet lifecycles.*** At the Cloud–Edge and Data Layer, safety must extend beyond an individual robot to model registries, training data, OTA updates, telemetry, shared memory, and fleet learning. Future research should develop verifiable data and model provenance, update attestation, staged deployment, rollback mechanisms, and quarantine procedures for compromised agents. Fleet-level safety also requires studying how poisoned maps, backdoored models, or malicious updates propagate among heterogeneous robots. Remote attestation and Byzantine-resilient collaboration should therefore be evaluated together with their communication, privacy, and availability costs.[237]

### Cross-layer challenges

***From static safety rules to runtime assurance and recovery.*** The previous sections also suggest that pre-deployment safety certification alone is insufficient. Embodied systems operate in open, partially observed, and changing environments. Even a model that passes offline benchmarks may encounter unseen objects, adversarial conditions, degraded sensors, or conflicting human instructions after deployment. This is why recoverability should be treated as an independent safety attribute, rather than being hidden under generic robustness. The survey defines embodied AI safety through controllability, interpretability, recoverability, and alignment, and highlights that physical actions are often irreversible, making post-failure compensation qualitatively different from pre-failure robustness.

Future work should build runtime assurance frameworks that continuously monitor whether the system is still operating inside a verified safety envelope. When uncertainty increases, the system should slow down, ask for clarification, switch to a conservative controller, isolate suspicious sensors, or trigger a hardware-level safe stop. This points toward a layered safety architecture in which large models provide flexible reasoning, verified monitors enforce hard constraints, and hardware mechanisms serve as the final containment boundary.[202,206-208] The long-term goal is not to make embodied agents "never fail," but to make them fail safely, visibly, and recoverably.

***From single-agent safety to multi-agent and fleet-level trust.*** Many future embodied systems will not operate as isolated robots. They will be deployed as fleets of service robots, warehouse robots, drones, autonomous vehicles, or household agents connected through cloud–edge infrastructures. This creates a new class of safety problems: an attack on one agent may propagate to others through shared maps, federated learning, OTA updates, common middleware, or coordinated task planning. The current four-layer architecture already includes cloud–edge data pipelines, model registries, telemetry, federated learning, and privacy-sensitive data flows as part of the embodied system stack.

Future research should therefore extend safety analysis from individual embodied agents to collective embodied ecosystems. Important problems include fleet-level remote attestation, trust management among heterogeneous robots, Byzantine-resilient collaborative perception, secure shared memory, and quarantine mechanisms for compromised agents.[184,197-199] A particularly important direction is to study how local failures become collective failures: for example, whether a poisoned map, compromised robot, or malicious update can induce synchronized unsafe behavior across a fleet. Addressing these challenges may draw on complementary techniques from robotics, distributed systems security, secure provenance mechanisms, federated learning security, and cyber-physical resilience, with the appropriate combination depending on the deployment context and threat model.

***From benchmarking isolated tasks to evaluating closed-loop consequences.*** Another major gap lies in evaluation. Existing benchmarks often measure whether a model answers safely, whether an agent refuses malicious instructions, or whether a perception model remains accurate under perturbation. These are useful, but they do not fully capture embodied risk. In embodied AI, the key question is not only whether the model made a wrong prediction, but whether that wrong prediction changed a plan, whether the plan changed a control command, and whether the command caused physical harm. The survey notes that evaluation should apply across all layers and all safety properties, rather than being confined to a single-layer benchmark.

Future benchmarks should therefore evaluate closed-loop consequences. A strong embodied safety benchmark should include perception perturbations, middleware failures, model-level jailbreaks, data poisoning, physical constraints, human proximity, and recovery outcomes in the same evaluation environment.[220–221,224–225] It should measure not only attack success rate, but also hazard severity, time-to-detection, recoverability, physical damage potential, human intervention burden, and generalization across embodiments. Simulation will be important, but purely simulated results are not enough; future benchmarks need calibrated digital twins, hardware-in-the-loop testing, and carefully bounded real-world validation.

***From human oversight to scalable safety governance.*** Human oversight remains necessary, but embodied AI makes oversight harder. The operator may not observe every sensor stream, understand every intermediate plan, or intervene before a fast physical action occurs. This creates a mismatch between the speed of physical execution and the speed of human judgment. Future safety governance should therefore focus on making embodied systems inspectable, interruptible, and accountable. Interpretability should not only explain a model's answer after the fact; it should expose task intent, environmental assumptions, uncertainty, planned trajectories, safety constraints, and possible failure modes before action execution.

This also means that future governance frameworks must bridge technical mechanisms and operational procedures. Safety cases, audit logs, model cards, red-team reports, runtime telemetry, and incident replay tools should be connected into a unified lifecycle.[66,79,202,206]. For high-risk embodied systems, deployment should not be treated as the end of evaluation, but as the beginning of continuous assurance. This direction aligns with the survey's broader Identification–Mitigation–Evaluation logic, where identifying risks, mitigating them, and evaluating residual risk form a continuous cycle rather than a one-time checklist.

## CONCLUSION

Safe deployment of embodied AI calls for an integrated approach to system safety and security. This survey examines the threat landscape through a four-layer reference architecture comprising the Physical Interaction Layer, System and Middleware Layer, Algorithm and Decision Layer, and Cloud–Edge and Data Layer. The reviewed literature illustrates how faults and attacks can propagate across layers through data dependencies and cyber-physical interactions. Sensor spoofing or compromised over-the-air updates can affect downstream perception, decision-making, or control, potentially leading to unsafe physical behavior when protective mechanisms are insufficient.

These findings motivate coordinated defenses across software, hardware, and data. We advocate combining semantic and contextual safety assessment with physical constraint enforcement through runtime assurance and hardware safeguards appropriate to each platform. These mechanisms can help block unsafe commands and mitigate hazards during execution, with their effectiveness depending on safety specifications, implementation, and operating conditions. We further advocate interactive, closed-loop benchmarking of multimodal and cross-layer risks, using explicit metrics for controllability, interpretability, recoverability, and alignment. Such evaluations can characterize system capabilities and limitations under defined conditions and inform safer deployment in environments shared with humans.

## REFERENCES

1. Duan J., Yu S., Tan H. L., et al. (2022). A survey of embodied AI: From simulators to research tasks. *IEEE Trans. Emerg. Top. Comput. Intell.* **6**:230–244. DOI:10.1109/tetci.2022.3141105
2. Liu Y., Chen W., Bai Y., et al. (2025). Aligning cyber space with physical world: A comprehensive survey on embodied AI. *IEEE/ASME Trans. Mechatron.* **30**:7253–7274. DOI:10.1109/TMECH.2025.3574943
3. Xing W., Li M., Li M., et al. (2026). Towards robust and secure embodied AI: A survey on vulnerabilities and attacks. *ACM Comput. Surv.* **58**:312. DOI:10.1145/3806048
4. Humayed A., Lin J., Li F., et al. (2017). Cyber-physical systems security—A survey. *IEEE Internet Things J.* **4**:1802–1831. DOI:10.1109/jiot.2017.2703172
5. Yaacoub J. P. A., Noura H. N., Salman O., et al. (2022). Robotics cyber security: vulnerabilities, attacks, countermeasures, and recommendations. *Int. J. Inf. Secur.* **21**:115–158. DOI:10.1007/s10207-021-00545-8
6. Ouyang L., Wu J., Jiang X., et al. (2022). Training language models to follow instructions with human feedback. *Adv. Neural Inf. Process. Syst.* **35**:27730–27744. DOI:10.52202/068431-2011
7. Rafailov R., Sharma A., Mitchell E., et al. (2023). Direct preference optimization: Your language model is secretly a reward model. *Adv. Neural Inf. Process. Syst.* **36**:53728–53741. DOI:10.52202/075280-2338
8. Robey A., Ravichandran Z., Kumar V., et al. (2025). Jailbreaking LLM-controlled robots. *2025 IEEE international conference on robotics and automation (ICRA)* (*IEEE*): 11948–11956. DOI: 10.1109/ICRA55743.2025.11128119
9. Ying Z., Wang L., Xiao Y., et al. (2026). Agentsafe: Benchmarking the safety of embodied agents on hazardous instructions. *Proceedings of the IEEE/CVF conference on computer vision and pattern recognition* 37664–37673.
10. Yin S., Pang X., Ding Y., et al. (2024). SafeAgentBench: A benchmark for safe task planning of embodied LLM agents. *arXiv.* DOI: 10.48550/arXiv.2412.13178
11. Xu Y., Han X., Deng G., et al. (2023). SoK: Rethinking sensor spoofing attacks against robotic vehicles from a systematic view. *2023 IEEE 8th European Symposium on Security and Privacy (EuroS&P)* 1082–1100. DOI:10.1109/eurosp57164.2023.00067
12. Cao Y., Xiao C., Cyr B., et al. (2019). Adversarial Sensor Attack on LiDAR-based Perception in Autonomous Driving. *Proceedings of the 2019 ACM SIGSAC conference on computer and communications security (CCS)* 2267–2281. DOI:10.1145/3319535.3339815
13. Botta A., Rotbei S., Zinno S., et al. (2023). Cyber security of robots: A comprehensive survey. *Intell. Syst. Appl.* **18**:200237. DOI:10.1016/j.iswa.2023.200237
14. Cárdenas A. A., Amin S. and Sastry S. (2008). Research challenges for the security of control systems. *3rd USENIX Workshop on Hot Topics in Security (HotSec 08)* (*USENIX Association*). Available at: https://www.usenix.org/conference/hotsec-08/research-challenges-security-control-systems
15. Stellios I., Kotzanikolaou P. and Grigoriadis C. (2021). Assessing IoT enabled cyber-physical attack paths against critical systems. *Comput. Secur.* **107**:102316. DOI:10.1016/j.cose.2021.102316
16. Liu T., Tian J., Wang J., et al. (2019). Integrated security threats and defense of cyber-physical systems. *Acta Autom. Sin.* **45**:5–24.
17. Kim J., Chen W., Soleymanzadeh D., et al. (2026). Modular safety guardrails are necessary for foundation-model-enabled robots in the real world. *arXiv.* DOI: 10.48550/arXiv.2602.04056
18. Lasota P. A., Fong T. and Shah J. A. (2017). A survey of methods for safe human-robot interaction. *Found. Trends Robot.* **5**:261–349. DOI:10.1561/2300000052
19. Desai A., Ghosh S., Seshia S. A., et al. (2019). SOTER: A runtime assurance framework for programming safe robotics systems. *49th Annual IEEE/IFIP International Conference on Dependable Systems and Networks*: 138–150. DOI: 10.1109/dsn.2019.00027
20. Albus J. S., Huang H. M., Messina E., et al. (2002). 4D/RCS version 2.0: A reference model architecture for unmanned vehicle systems. (*National Institute of Standards and Technology*).
21. Ahmad A. and Babar M. A. (2016). Software architectures for robotic systems: A systematic mapping study. *J. Syst. Softw.* **122**:16–39. DOI:10.1016/j.jss.2016.08.039
22. Elkady A. and Sobh T. (2012). Robotics middleware: A comprehensive literature survey and attribute-based bibliography. *J. Robot.* **2012**:959013. DOI:10.1155/2012/959013
23. Kehoe B., Patil S., Abbeel P., et al. (2015). A survey of research on cloud robotics and automation. *IEEE Trans. Autom. Sci. Eng.* **12**:398–409. DOI:10.1109/tase.2014.2376492
24. Hu G., Tay W. P. and Wen Y. (2012). Cloud robotics: Architecture, challenges and applications. *IEEE Network* **26**:21–28. DOI:10.1109/mnet.2012.6201212
25. De Santis A., Siciliano B., De Luca A., et al. (2008). An atlas of physical human–robot interaction. *Mech. Mach. Theory* **43**:253–270. DOI:10.1016/j.mechmachtheory.2007.03.003
26. Quigley M., Gerkey B. P., Conley K., et al. (2009). ROS: An open-source robot operating system. *Open-source software workshop of the international conference on robotics and automation (ICRA).*
27. Dieber B., Breiling B., Taurer S., et al. (2017). Security for the robot operating system. *Robot. Auton. Syst.* **98**:192–203. DOI:10.1016/j.robot.2017.09.017
28. Ahn M., Brohan A., Brown N., et al. (2023). Do as I can, not as I say: Grounding language in robotic affordances. *Proceedings of the 6th Conference on Robot Learning (CoRL)* 287–318.
29. Driess D., Xia F., Sajjadi M. S. M., et al. (2023). PaLM-E: An embodied multimodal language model. *Proceedings of the 40th International Conference on Machine Learning* (*PMLR*), Proceedings of Machine Learning Research 202: 8469–8488. Available at: https://proceedings.mlr.press/v202/driess23a.html
30. Brohan A., Brown N., Carbajal J., et al. (2023). RT-2: Vision-language-action models transfer web knowledge to robotic control. *Proceedings of the 7th Conference on Robot Learning (CoRL)*: 2165–2183.
31. Kim M. J., Pertsch K., Karamcheti S., et al. (2025). OpenVLA: An open-source vision-language-action model. *Proceedings of the 8th Conference on Robot Learning (CoRL)* (*PMLR*), Proceedings of Machine Learning Research 270: 2679–2713.
32. Black K., Brown N., Driess D., et al. (2024). π0: A vision-language-action flow model for general robot control. *arXiv.* DOI: 10.48550/arXiv.2410.24164
33. NVIDIA, Bjorck J., Castañeda F., et al. (2025). GR00T N1: An open foundation model for generalist humanoid robots. *arXiv.* DOI: 10.48550/arXiv.2503.14734
34. Open X-Embodiment Collaboration, O'Neill A., Rehman A., et al. (2024). Open X-Embodiment: Robotic learning datasets and RT-X models. *2024 IEEE international conference on robotics and automation (ICRA)*: 6892–6903.
35. Kim K., Kim J. S., Jeong S., et al. (2021). Cybersecurity for autonomous vehicles: Review of attacks and defense. *Comput. Secur.* **103**:102150. DOI:10.1016/j.cose.2020.102150
36. Sun X., Yu F. R. and Zhang P. (2022). A survey on cyber-security of connected and autonomous vehicles (CAVs). *IEEE Trans. Intell. Transp. Syst.* **23**:6240–6259. DOI:10.

1109/tits.2021.3085297
37. Giraldo J., Urbina D., Cardenas A., et al. (2018). A survey of physics-based attack detection in cyber-physical systems. *ACM Comput. Surv.* **51**:1−36. DOI:10.1145/3203245
38. International Organization for Standardization (2024). ISO/IEC TR 5469: 2024 — Artificial intelligence — Functional safety and AI systems. (*International Organization for Standardization*). Available at: https://www.iso.org/standard/81283.html
39. Ames A. D., Xu X., Grizzle J. W., et al. (2017). Control barrier function based quadratic programs for safety critical systems. *IEEE Trans. Autom. Control* **62**:3861−3876. DOI:10.1109/tac.2016.2638961
40. Alshiekh M., Bloem R., Ehlers R., et al. (2018). Safe reinforcement learning via shielding. *Proceedings of the 32nd AAAI Conference on Artificial Intelligence (AAAI)* **32**:2669−2678. DOI:10.1609/aaai.v32i1.11797
41. Anjomshoae S., Najjar A., Calvaresi D., et al. (2019). Explainable agents and robots: Results from a systematic literature review. *Proceedings of the 18th International Conference on Autonomous Agents and MultiAgent Systems (AAMAS)* 1078−1088. DOI:10.65109/kczb5817
42. Sakai T. and Nagai T. (2022). Explainable autonomous robots: A survey and perspective. *Adv. Robot.* **36**:219−238. DOI:10.1080/01691864.2022.2029720
43. Visinsky M. L., Cavallaro J. R. and Walker I. D. (1994). Robotic fault detection and fault tolerance: A survey. *Reliab. Eng. Syst. Saf.* **46**:139−158. DOI:10.1016/0951-8320(94)90132-5
44. Leucker M. and Schallhart C. (2009). A brief account of runtime verification. *J. Log. Algebr. Program.* **78**:293−303. DOI:10.1016/j.jlap.2008.08.004
45. Amodei D., Olah C., Steinhardt J., et al. (2016). Concrete problems in AI safety. *arXiv.* DOI: 10.48550/arXiv.1606.06565
46. Ji J., Qiu T., Chen B., et al. (2023). AI alignment: A comprehensive survey. *arXiv.* DOI: 10.48550/arXiv.2310.19852
47. International Electrotechnical Commission (2010). IEC 61508: Functional safety of electrical/electronic/programmable electronic safety-related systems, Edition 2.0. (*International Electrotechnical Commission*). Available at: https://webstore.iec.ch/en/publication/22273
48. International Organization for Standardization (2011). ISO 10218-1: 2011 and ISO 10218-2: 2011 — Robots and robotic devices — Safety requirements for industrial robots. (*International Organization for Standardization*).
49. Leveson N. G. (2012). Engineering a safer world: Systems thinking applied to safety. (*MIT Press*).
50. International Organization for Standardization (2016). ISO/TS 15066: 2016 — Robots and robotic devices — Collaborative robots. (*International Organization for Standardization*). Available at: https://www.iso.org/standard/62996.html
51. Haddadin S., Albu-Schäffer A. and Hirzinger G. (2009). Requirements for safe robots: Measurements, analysis and new insights. *Int. J. Robot. Res.* **28**:1507−1527. DOI:10.1177/0278364909343970
52. Villani V., Pini F., Leali F., et al. (2018). Survey on human–robot collaboration in industrial settings: Safety, intuitive interfaces and applications. *Mechatronics* **55**:248−266. DOI:10.1016/j.mechatronics.2018.02.009
53. Underwriters Laboratories (2023). UL 4600: Standard for safety for the evaluation of autonomous products, edition 3. (*Underwriters Laboratories*). Available at: https://www.ul.com/news/ul-4600-edition-3-updates-incorporate-autonomous-trucking
54. Zhang H., Zhu C., Wang X., et al. (2025). BadRobot: Jailbreaking embodied LLM agents in the physical world. *The thirteenth international conference on learning representations (ICLR).*
55. Ruan Y., Dong H., Wang A., et al. (2024). Identifying the risks of LM agents with an LM-emulated sandbox. *Proceedings of the 12th International Conference on Learning Representations (ICLR).*
56. Li H., Wei M., Huang J., et al. (2019). Survey on cyber-physical system technologies. *Acta Autom. Sin.* 45: 37−50.
57. Buldyrev S. V., Parshani R., Paul G., et al. (2010). Catastrophic cascade of failures in interdependent networks. *Nature* **464**:1025−1028. DOI:10.1038/nature08932
58. Shin H., Kim D., Kwon Y., et al. (2017). Illusion and Dazzle: Adversarial Optical Channel Exploits against LiDARs for Automotive Applications. *Cryptographic hardware and embedded systems (CHES)* 445−467. DOI:10.1007/978-3-319-66787-4_22
59. Son Y., Shin H., Kim D., et al. (2015). Rocking drones with intentional sound noise on gyroscopic sensors. *Proceedings of the 24th USENIX Security Symposium* 881−896.
60. Quarta D., Pogliani M., Polino M., et al. (2017). An experimental security analysis of an industrial robot controller. *2017 IEEE symposium on security and privacy (s&p)* 268−286. DOI:10.1109/sp.2017.20
61. Schneier B. (1999). Attack trees: Modeling security threats. *Dr. Dobb's J.* **24**:21−29.
62. Shostack A. (2014). Threat modeling: Designing for security. (*Wiley*).
63. UcedaVelez T. and Morana M. M. (2015). Risk centric threat modeling: Process for attack simulation and threat analysis. (*Wiley*).
64. International Organization for Standardization and Society of Automotive Engineers (2021). ISO/SAE 21434: 2021 — road vehicles — cybersecurity engineering. (*International Organization for Standardization and SAE International*). Available at: https://www.iso.org/standard/70918.html
65. National Institute of Standards and Technology (2024). The NIST Cybersecurity Framework (CSF) 2.0. (*National Institute of Standards and Technology*). Available at: https://www.nist.gov/publications/nist-cybersecurity-framework-csf-20
66. Tabassi E. (2023). Artificial intelligence risk management framework (AI RMF 1.0). (*National Institute of Standards and Technology*). DOI: 10.6028/NIST.AI.100-1
67. El-Rewini Z., Sadatsharan K., Selvaraj D. F., et al. (2020). Cybersecurity challenges in vehicular communications. *Veh. Commun.* **23**:100214. DOI:10.1016/j.vehcom.2019.100214
68. Haskard A. and Herath D. (2025). Secure robotics: Navigating challenges at the nexus of safety, trust, and cybersecurity in cyber-physical systems. *ACM Comput. Surv.* **57**:1−48. DOI:10.1145/3723050
69. Zacharaki A., Kostavelis I., Gasteratos A., et al. (2020). Safety bounds in human−robot interaction: A survey. *Saf. Sci.* **127**:104667. DOI:10.1016/j.ssci.2020.104667
70. Dibaji S. M., Pirani M., Flamholz D. B., et al. (2019). A systems and control perspective of CPS security. *Annu. Rev. Control* **47**:394−411. DOI:10.1016/j.arcontrol.2019.04.011
71. Petit J. and Shladover S. E. (2015). Potential cyberattacks on automated vehicles. *IEEE Trans. Intell. Transp. Syst.* **16**:546−556. DOI:10.1109/tits.2014.2342271
72. Guesmi A., Hanif M. A., Ouni B., et al. (2024). Physical adversarial attacks for camera-based smart systems: Current trends, categorization, applications, research challenges, and future outlook. *IEEE Access* **12**:14150−14174. DOI:10.1109/access.2023.3321118
73. Girdhar M., Hong J. and Moore J. (2023). Cybersecurity of autonomous vehicles: A systematic literature review of adversarial attacks and defense models. *IEEE Open J. Veh. Technol.* **4**:417−437. DOI:10.1109/ojvt.2023.3265363
74. Deng Z., Guo Y., Han C., et al. (2025). AI agents under threat: A survey of key security challenges and future pathways. *ACM Comput. Surv.* **57**:182. DOI:10.1145/3716628
75. He F., Zhu T., Ye D., et al. (2025). The emerged security and privacy of LLM agent: A survey with case studies. *ACM Comput. Surv.* **58**:1−36. DOI:10.1145/3773080
76. Liu Y., Yao Y., Ton J. F., et al. (2023). Trustworthy LLMs: A survey and guideline for evaluating large language models' alignment. *NeurIPS 2023 workshop on socially responsible language modelling research (SoLaR).*
77. Neupane S., Mitra S., Fernandez I. A., et al. (2024). Security considerations in AI-robotics: A survey of current methods, challenges, and opportunities. *IEEE Access* **12**:22072−22097. DOI:10.1109/access.2024.3363657
78. Wang Z., Hu J. and Mu R. (2025). Safety of embodied navigation: A survey. *Proceedings of the 34th International Joint Conference on Artificial Intelligence (IJCAI) — Survey Track* 10714−10722. DOI:10.24963/ijcai.2025/1189
79. Xu W., Ji X., Yan C., et al. (2025). Embodied artificial intelligence security and governance. *Bull. Chin. Acad. Sci.* **40**:429−439. DOI:10.16418/j.issn.1000-3045.20250218002
80. Lu X., Huang Z., Li X., et al. (2024). POEX: Towards policy executable jailbreak attacks against the LLM-based robots. *arXiv.* DOI: 10.48550/arXiv.2412.16633
81. Tang L., Wang R., Liu Z., et al. (2024). Scenario-based accelerated testing for SOTIF in autonomous driving: a review. *IEEE Internet Things J.* **12**:1453−1470. DOI:10.1109/jiot.2024.3490598
82. Zhang X., Dong H., Zhang H., et al. (2025). A real-time, robust and versatile visual-SLAM framework based on deep learning networks. *IEEE Trans. Instrum. Meas.* **74**:1−13. DOI:10.1109/tim.2025.3527618
83. Chakraborty A., Alam M., Dey V., et al. (2021). A survey on adversarial attacks and defences. *CAAI Trans. Intell. Technol.* **6**:25−45. DOI:10.1049/cit2.12028
84. Liu J., Levine A., Lau C. P., et al. (2022). Segment and complete: Defending object detectors against adversarial patch attacks with robust patch detection. *Proceedings of the IEEE/CVF conference on computer vision and pattern recognition* 14973−14982. DOI:10.1109/cvpr52688.2022.01455
85. Nassi B., Mirsky Y., Nassi D., et al. (2020). Phantom of the ADAS: Securing advanced driver-assistance systems from split-second phantom attacks. *ACM conference on computer and communications security (CCS)* 293−308. DOI:10.1145/3372297.3423359
86. Li Y., Yang F., Liu Q., et al. (2023). Light can be dangerous: Stealthy and effective physical-world adversarial attack by spot light. *Comput. Secur.* **132**:103345. DOI:10.1016/j.cose.2023.103345
87. Cheng Y., Ji X., Zhu W., et al. (2023). Adversarial computer vision via acoustic manipulation of camera sensors. *IEEE Trans. Dependable Secure Comput.* **21**:3734−3750. DOI:10.1109/tdsc.2023.3334618
88. Duan R., Mao X., Qin A. K., et al. (2021). Adversarial laser beam: Effective physical-world attack to DNNs in a blink. *Proceedings of the IEEE/CVF conference on computer vision and pattern recognition (CVPR)* 16057−16066. DOI:10.1109/cvpr46437.2021.01580
89. Yan C., Xu Z., Yin Z., et al. (2022). Rolling colors: Adversarial laser exploits against traffic light recognition. *arXiv.* DOI: 10.48550/arXiv.2204.02675
90. Cheng Z., Liang J., Choi H., et al. (2022). Physical attack on monocular depth estimation with optimal adversarial patches. *Computer vision – ECCV 2022* (*Springer*), Lecture Notes in Computer Science *13698*: 514−532. DOI: 10.1007/978-3-031-19839-7_30
91. Jing P., Tang Q., Du Y., et al. (2021). Too good to be safe: Tricking lane detection in autonomous driving with crafted perturbations. *30th USENIX Security Symposium*

*(USENIX Security 21)* 3237–3254.
92. Jia Y., Lu Y., Shen J., et al. (2020). Fooling Detection Alone is Not Enough: Adversarial Attack against Multiple Object Tracking. *International conference on learning representations (ICLR).*
93. Davidson D., Wu H., Jellinek R., et al. (2016). Controlling UAVs with sensor input spoofing attacks. *Workshop on offensive technologies (WOOT).*
94. Burrus N., Abderrahim M., Garcia J., et al. (2011). Object Reconstruction and Recognition leveraging an RGB-D camera. *Proceedings of the 12th IAPR Conference on Machine Vision Applications*: 3–6.
95. Li Y. and Ibanez-Guzman J. (2020). LiDAR for autonomous driving: The principles, challenges, and trends for automotive LiDAR and perception systems. *IEEE Signal Process. Mag.* 37:50–61. DOI:10.1109/msp.2020.2973615
96. Kim B. K. and Sumi Y. (2020). Vision-based safety-related sensors in low visibility by fog. *Sensors 20*(10): 2812. DOI: 10.3390/s20102812
97. Petit J., Stottelaar B., Feiri M., et al. (2015). Remote attacks on automated vehicles sensors: Experiments on camera and LiDAR. *Black hat europe.*
98. Jin Z., Ji X., Cheng Y., et al. (2023). PLA-LiDAR: Physical laser attacks against LiDAR-based 3D object detection in autonomous vehicle. *IEEE symposium on security and privacy (s&p)* 1822–1839. DOI:10.1109/sp46215.2023.10179458
99. Ahmad N., Ghazilla R. A. R., Khairi N. M., et al. (2013). Reviews on various inertial measurement unit (IMU) sensor applications. *IJSPS* 1:256–262. DOI:10.12720/ijsps.1.2.256-262
100. Burnett K., Schoellig A. P. and Barfoot T. D. (2025). IMU as an input versus a measurement of the state in inertial-aided state estimation. *Robotica* 43:1–21. DOI:10.1017/s0263574724002121
101. Wu Y., Kuang J., Niu X., et al. (2025). Wheel-GINS: A GNSS/INS integrated navigation system with a wheel-mounted IMU. *IEEE Trans. Intell. Transp. Syst.* 26:6891–6903. DOI:10.1109/tits.2025.3527815
102. Gao M., Zhang L., Shen L., et al. (2023). Exploring practical acoustic transduction attacks on inertial sensors in MDOF systems. *IEEE Trans. Mob. Comput.* 23:3539–3557. DOI:10.1109/tmc.2023.3277287
103. Wang Z., Wang K., Yang B., et al. (2017). Sonic gun to smart devices. *Black hat USA.*
104. Ji X., Cheng Y., Zhang Y., et al. (2021). Poltergeist: Acoustic adversarial machine learning against cameras and computer vision. *IEEE symposium on security and privacy (s&p)* 160–175. DOI:10.1109/sp40001.2021.00091
105. Dey V., Pudi V., Chattopadhyay A., et al. (2018). Security vulnerabilities of unmanned aerial vehicles and countermeasures: An experimental study. *International conference on VLSI design* 398–403. DOI:10.1109/vlsid.2018.97
106. John A. Volpe National Transportation Systems Center (2001). Vulnerability assessment of the transportation infrastructure relying on the Global Positioning System. (*U.S. Department of Transportation, Office of the Assistant Secretary for Transportation Policy*). Available at: https://rosap.ntl.bts.gov/view/dot/8435
107. Tippenhauer N. O., Pöpper C., Rasmussen K. B., et al. (2011). On the requirements for successful GPS spoofing attacks. *Proceedings of the 18th ACM Conference on Computer and Communications Security* 75–86. DOI:10.1145/2046707.2046719
108. Zeng K., Liu S., Shu Y., et al. (2018). All your GPS are belong to us: Towards stealthy manipulation of road navigation systems. *27th USENIX Security Symposium (USENIX Security 18)*: 1527–1544.
109. Nighswander T., Ledvina B. M., Diamond J., et al. (2012). GPS software attacks. *ACM conference on computer and communications security (CCS)* 450–461. DOI:10.1145/2382196.2382245
110. Shen J., Won J. Y., Chen Z., et al. (2020). Drift with devil: Security of Multi-Sensor fusion based localization in High-Level autonomous driving under GPS spoofing. *29th USENIX Security Symposium (USENIX Security 20)* (*USENIX Association*): 931–948.
111. Li Y., Wen C., Juefei-Xu F., et al. (2021). Fooling LiDAR perception via adversarial trajectory perturbation. *IEEE international conference on computer vision* 7898–7907. DOI:10.1109/iccv48922.2021.00780
112. Tang K., Shen J. and Chen Q. A. (2021). Fooling perception via location: A case of region-of-interest attacks on traffic light detection in autonomous driving. *NDSS workshop on automotive and autonomous vehicle security (AutoSec).* DOI: 10.14722/autosec.2021.23029
113. Dahiya R. S., Metta G., Valle M., et al. (2009). Tactile sensing—from humans to humanoids. *IEEE Trans. Robot. 26*(1): 1–20. DOI: 10.1109/tro.2009.2033627
114. Jiang J. and Luo S. (2022). Robotic perception of object properties using tactile sensing. *Tactile sensing, skill learning, and robotic dexterous manipulation* (*Elsevier*), pp. 23–44. DOI: 10.1016/b978-0-32-390445-2.00009-x
115. Mittendorfer P. and Cheng G. (2011). Humanoid multimodal tactile-sensing modules. *IEEE Trans. Robot. 27*(3): 401–410. DOI: 10.1109/tro.2011.2106330
116. Tu Y., Rampazzi S. and Hei X. (2022). Towards adversarial control loops in sensor attacks: a case study to control the kinematics and actuation of embedded systems. *arXiv.* DOI: 10.48550/arXiv.2203.07670
117. Shi J., Dai Y., Cheng Y., et al. (2023). Embedment of sensing elements for robust, highly sensitive, and cross-talk–free iontronic skins for robotics applications. *Sci. Adv.* 9:eadf8831. DOI:10.1126/sciadv.adf8831
118. Regenscheid A. (2018). Platform firmware resiliency guidelines. (*National Institute of Standards and Technology (NIST)*). DOI: 10.6028/nist.sp.800-193
119. Gandolfi K., Mourtel C. and Olivier F. (2001). Electromagnetic analysis: Concrete results. *Cryptographic hardware and embedded systems—CHES 2001: Third international workshop paris, france, may 14–16, 2001 proceedings 3*: 251–261. DOI: 10.1007/3-540-44709-1_21
120. Kocher P., Jaffe J. and Jun B. (1999). Differential power analysis. *Annual international cryptology conference (CRYPTO'99)* 388–397. DOI:10.1007/3-540-48405-1_25
121. Mangard S., Oswald E. and Popp T. (2008). Power analysis attacks: Revealing the secrets of smart cards. (*Springer*).
122. Khuat V., Danger J. L. and Dutertre J. M. (2021). Laser fault injection in a 32-bit micro-controller: from the flash interface to the execution pipeline. *2021 Workshop on fault detection and tolerance in cryptography (FDTC)*: 74–85. DOI: 10.1109/fdtc53659.2021.00020
123. Bhunia S., Hsiao M. S., Banga M., et al. (2014). Hardware Trojan attacks: Threat analysis and countermeasures. *Proc. IEEE* 102:1229–1247. DOI:10.1109/jproc.2014.2334493
124. Tehranipoor M. and Koushanfar F. (2010). A survey of hardware trojan taxonomy and detection. *IEEE Des. Test. Comput.* 27:10–25. DOI:10.1109/mdt.2010.7
125. Xiao K., Forte D., Jin Y., et al. (2016). Hardware trojans: Lessons learned after one decade of research. *ACM Trans. Des. Autom. Electron. Syst.* 22:1–23.
126. Barenghi A., Breveglieri L., Koren I., et al. (2012). Fault injection attacks on cryptographic devices: Theory, practice, and countermeasures. *Proc. IEEE* 100:3056–3076. DOI:10.1109/jproc.2012.2188769
127. Qiu P., Wang D., Lyu Y., et al. (2019). Voltjockey: Breaching TrustZone by software-controlled voltage manipulation over multi-core frequencies. *ACM SIGSAC conference on computer and communications security (CCS)*: 195–209.
128. Moro N., Dehbaoui A., Heydemann K., et al. (2013). Electromagnetic fault injection: towards a fault model on a 32-bit microcontroller. *2013 Workshop on fault diagnosis and tolerance in cryptography*: 77–88. DOI: 10.1109/fdtc.2013.9
129. Maggi F., Quarta D., Pogliani M., et al. (2017). Rogue robots: Testing the limits of an industrial robot's security. (*Trend Micro and Politecnico di Milano*). Available at: https://documents.trendmicro.com/assets/wp/wp-industrial-robot-security.pdf
130. Kasprzyczak L., Manowska A. and Dźwiarek M. (2025). Cybersecurity requirements for industrial machine control systems. *Appl. Sci.* 15:1267. DOI:10.3390/app15031267
131. Jiménez Naharro R., Gómez-Bravo F. and López de Ahumada Gutiérrez R. (2025). Exploring hardware vulnerabilities in robotic actuators: A case of man-in-the-middle attacks. *Electronics* 14:4909. DOI:10.3390/electronics14244909
132. Sabt M., Achemlal M. and Bouabdallah A. (2015). Trusted execution environment: What it is, and what it is not. *2015 IEEE trustcom/BigDataSE/ispa* 1:57–64. DOI:10.1109/trustcom.2015.357
133. Arbaugh W. A., Farber D. J. and Smith J. M. (1997). A secure and reliable bootstrap architecture. *Proceedings. 1997 IEEE symposium on security and privacy (cat. no. 97CB36097)* 65–71. DOI:10.1109/secpri.1997.601317
134. Suh G. E. and Devadas S. (2007). Physical unclonable functions for device authentication and secret key generation. *Proceedings of the 44th annual design automation conference* 9–14. DOI:10.1145/1278480.1278484
135. Haldar V., Chandra D. and Franz M. (2004). Semantic remote attestation: A virtual machine directed approach to trusted computing. *USENIX virtual machine research and technology symposium 2004.*
136. Ivanov R., Pajic M. and Lee I. (2016). Attack-resilient sensor fusion for safety-critical cyber-physical systems. *ACM Trans. Embed. Comput. Syst.* 15:1–24. DOI:10.1145/2847418
137. Clark S. S., Ransford B., Rahmati A., et al. (2013). WattsUpDoc: Power side channels to nonintrusively discover untargeted malware on embedded medical devices. *2013 USENIX workshop on health information technologies (HealthTech 13).*
138. Bao C., Forte D. and Srivastava A. (2015). Temperature tracking: Toward robust run-time detection of hardware Trojans. *IEEE Trans. Comput.-Aided Des. Integr. Circuits Syst.* 34:1577–1585. DOI:10.1109/tcad.2015.2424929
139. Pan M. C., Van Brussel H. and Sas P. (1998). Intelligent joint fault diagnosis of industrial robots. *Mech. Syst. Signal Process.* 12:571–588. DOI:10.1006/mssp.1997.0124
140. Guin U., Huang K., DiMase D., et al. (2014). Counterfeit integrated circuits: A rising threat in the global semiconductor supply chain. *Proc. IEEE* 102:1207–1228. DOI:10.1109/jproc.2014.2332291
141. Sun J., Cao Y., Chen Q. A., et al. (2020). Towards robust LiDAR-based perception in autonomous driving: General black-box adversarial sensor attack and countermeasures. *29th USENIX Security Symposium*: 877–894.
142. Szegedy C., Zaremba W., Sutskever I., et al. (2014). Intriguing properties of neural networks. *International conference on learning representations (ICLR)* 1–10.
143. Goodfellow I. J., Shlens J. and Szegedy C. (2015). Explaining and harnessing adversarial examples. *International conference on learning representations (ICLR)* 1–11.
144. Brown T. B., Mané D., Roy A., et al. (2017). Adversarial patch. *arXiv.* DOI: 10.48550/arXiv.1712.09665
145. Eykholt K., Evtimov I., Fernandes E., et al. (2018). Robust physical-world attacks on deep learning visual classification. *Proceedings of the IEEE/CVF conference on computer vision and pattern recognition* 1625–1634. DOI:10.1109/cvpr.2018.00175
146. Eykholt K., Evtimov I., Fernandes E., et al. (2018). Physical adversarial examples for

object detectors. *12th USENIX Workshop on Offensive Technologies*.
147. Mehmood U., Sheikhi S., Bak S., et al. (2022). The black-box simplex architecture for runtime assurance of autonomous CPS. *NASA formal methods* 231–250. DOI:10.1007/978-3-031-06773-0_12
148. García J. and Fernández F. (2015). A comprehensive survey on safe reinforcement learning. *J. Mach. Learn. Res.* **16**:1437–1480.
149. Achiam J., Held D., Tamar A., et al. (2017). Constrained policy optimization. *Proceedings of the 34th International Conference on Machine Learning*: 22–31.
150. Dalal G., Dvijotham K., Vecerik M., et al. (2018). Safe exploration in continuous action spaces. *arXiv*. DOI: 10.48550/arXiv.1801.08757
151. Ames A. D., Coogan S., Egerstedt M., et al. (2019). Control barrier functions: Theory and applications. *2019 18th European Control Conference* 3420–3431. DOI:10.23919/ecc.2019.8796030
152. Moulard T., Hortala J., Perez X., et al. (2019). ROS 2 Robotic Systems Threat Model. Available at: https://design.ros2.org/articles/ros2_threat_model.html
153. Open Robotics (2020). ROS 2 security enclaves. (*Open Robotics*). Available at: https://design.ros2.org/articles/ros2_security_enclaves.html
154. DiLuoffo V., Michalson W. R. and Sunar B. (2018). Robot operating system 2: The need for a holistic security approach to robotic architectures. *Int. J. Adv. Robot. Syst.* **15**:1–15. DOI:10.1177/1729881418770011
155. Deng G., Xu G., Zhou Y., et al. (2022). On the (In)Security of secure ROS2. *Proceedings of the 2022 ACM SIGSAC conference on computer and communications security* 739–753. DOI:10.1145/3548606.3560681
156. Pandya N. V., Kumar H., Pillai G. M., et al. (2024). Decentralized information-flow control for ROS2. *31st Annual Network and Distributed System Security Symposium (NDSS 2024)* (*Internet Society*): 1–18. DOI: 10.14722/ndss.2024.24101
157. Jiao R., Xie S., Yue J., et al. (2025). Can we trust embodied agents? Exploring backdoor attacks against embodied LLM-based decision-making systems. *The thirteenth international conference on learning representations (ICLR 2025)* (*ICLR*). DOI: 10.48550/arXiv.2405.20774
158. OWASP Foundation (2025). OWASP top 10 for large language model applications. (*OWASP Foundation*). Available at: https://genai.owasp.org/resource/owasp-top-10-for-llm-applications-2025/
159. MITRE (2026). MITRE ATLAS: Adversarial threat landscape for artificial-intelligence systems. (*MITRE*). Available at: https://atlas.mitre.org/
160. European Parliament and Council of the European Union (2024). Regulation (EU) 2024/1689: Artificial intelligence act. (*Publications Office of the European Union*). Available at: https://eur-lex.europa.eu/eli/reg/2024/1689/oj
161. International Society of Automation (2026). ISA/IEC 62443 series of standards: Security for industrial automation and control systems. (*International Society of Automation*). Available at: https://www.isa.org/standards-and-publications/isa-standards/isa-iec-62443-series-of-standards
162. Hu H., Salcic Z., Sun L., et al. (2022). Membership inference attacks on machine learning: A survey. *ACM Comput. Surv.* **54**:1–37. DOI:10.1145/3523273
163. Li M., Ding W. and Zhao D. (2024). Privacy risks in reinforcement learning for household robots. *IEEE international conference on robotics and automation (ICRA)* 5148–5154. DOI:10.1109/icra57147.2024.10610832
164. Dietrich M., Krüger M. and Weisswange T. H. (2023). What should a robot disclose about me? A study about privacy-appropriate behaviors for social robots. *Front. Robot. AI* **10**:1236733. DOI:10.3389/frobt.2023.1236733
165. Stapels J. G., Penner A., Diekmann N., et al. (2023). Never trust anything that can think for itself if you can't control its privacy settings: The influence of a robot's privacy settings on users' attitudes and willingness to self-disclose. *Int. J. Soc. Robot.* **15**:1487–1505. DOI:10.1007/s12369-023-01043-8
166. Su H., Cui Y., Santina C. D., et al. (2026). Embodied AI: Bridging robotics and AI toward real-world applications [from the guest editors]. *IEEE Robot. Autom. Mag.* **33**:6–7. DOI:10.1109/mra.2026.3653534
167. Nazarczuk M., Behrens J. K., Stepanova K., et al. (2025). Closed loop interactive embodied reasoning for robot manipulation. *2025 IEEE international conference on robotics and automation (ICRA)* 13722–13729. DOI:10.1109/icra55743.2025.11127480
168. Gurdur Broo D. (2026). Physical AI in cyber-physical systems: from digital to embodied industrial agents. *J. Ind. Inf. Integr.* **49**:101038. DOI:10.1016/j.jii.2025.101038
169. Firoozi R., Tucker J., Tian S., et al. (2025). Foundation models in robotics: Applications, challenges, and the future. *Int. J. Robot. Res.* **44**:701–739. DOI:10.1177/02783649241281508
170. Hu T., Gong Z., Kong L., et al. (2026). NavThinker: Action-conditioned world models for coupled prediction and planning in social navigation. *arXiv*. DOI: 10.48550/arXiv.2603.15359
171. Gemini Robotics Team, Abeyruwan S., Ainslie J., et al. (2025). Gemini robotics: Bringing AI into the physical world. *arXiv*. DOI: 10.48550/arXiv.2503.20020
172. Ma Y., Song Z., Zhuang Y., et al. (2024). A survey on vision-language-action models for embodied AI. *arXiv*. DOI: 10.48550/arXiv.2405.14093
173. Kojima T., Zhu Y., Iwasawa Y., et al. (2025). A comprehensive survey on physical risk control in the era of foundation model-enabled robotics. *arXiv*. DOI: 10.48550/arXiv.2505.12583
174. Xing W. and Shen J. (2024). Security control of cyber–physical systems under cyber attacks: A survey. *Sensors* **24**:3815. DOI:10.3390/s24123815
175. Deng Y., Wu T., Wu D., et al. (2026). Efficient coordination with the system-level shared state: An embodied-AI native modular framework. *arXiv*. DOI: 10.48550/arXiv.2601.13945
176. Xie M. and Wei-Kocsis J. (2026). From prompt to physical action: Structured backdoor attacks on LLM-mediated robotic control systems. *arXiv*. DOI: 10.48550/arXiv.2604.03890
177. Li A., Wang J., Baruah S., et al. (2024). An empirical study of performance interference: Timing violation patterns and impacts. *2024 IEEE 30th Real-Time and Embedded Technology and Applications Symposium (RTAS)* 320–333. DOI:10.1109/rtas61025.2024.00033
178. Hsiao Y. P., Li Y., Gamal Y., et al. (2026). Glitch in the sky: Exploiting voltage fault injection in UAV flight controllers. *arXiv*. DOI: 10.48550/arXiv.2604.16699
179. Banerjee D., Tan C. M. and Baruah N. A. (2024). Application of component failure physics for the reliability assessment of an autonomous braking system. *Sci. Rep.* **14**:28835. DOI:10.1038/s41598-024-80476-1
180. Liu Z., Shanmugam D. and Schaumont P. (2024). FaultDetective: Explainable to a fault, from the design layout to the software. *IACR Trans. Cryptogr. Hardw. Embed. Syst.* **2024**:610–632.
181. Liu J., Li H., Wang H., et al. (2025). TimeTravel: Real-time timing drift attack on system time using acoustic waves. *34th USENIX Security Symposium (USENIX Security 25)* 3885–3902.
182. Fan W., Lane J., Liu Q., et al. (2024). Incorporating System-level Safety Requirements in Perception Models via Reinforcement Learning. *arXiv*. DOI: 10.48550/arXiv.2412.02951
183. Huang Y., Hao Y., Yu B., et al. (2025). DaDu-Corki: Algorithm-architecture co-design for embodied AI-powered robotic manipulation. *Proceedings of the 52nd Annual International Symposium on Computer Architecture (ISCA '25)* (*ACM*): 327–343. DOI: 10.1145/3695053.3731099
184. Xie T., Qi Y., Wen J., et al. (2026). CREATE: Cross-layer resilience characterization and optimization for efficient yet reliable embodied AI systems. *Proceedings of the 31st ACM International Conference on Architectural Support for Programming Languages and Operating Systems, Volume 2*: 493–510.
185. Sato T., Suzuki R., Hayakawa Y., et al. (2025). On the realism of LiDAR spoofing attacks against autonomous driving vehicle at high speed and long distance. *Network and distributed system security symposium (NDSS)*. DOI: 10.14722/ndss.2025.230628
186. Sadeghi J., Lord N. A., Redford J., et al. (2023). Attacking motion planners using adversarial perception errors. *arXiv*. DOI: 10.48550/arXiv.2311.12722
187. Wu W., Pierazzi F., Du Y., et al. (2024). Characterizing physical adversarial attacks on robot motion planners. *2024 IEEE international conference on robotics and automation (ICRA)* 14319–14325. DOI:10.1109/icra57147.2024.10610344
188. Mayoral-Vilches V., Carbajo U. A. and Gil-Uriarte E. (2020). Industrial robot ransomware: Akerbeltz. *2020 Fourth IEEE international conference on robotic computing (IRC)* 432–435. DOI:10.1109/irc.2020.00080
189. Pu H., He L., Cheng P., et al. (2024). CORMAND2: A deception attack against industrial robots. *Engineering* **32**:186–201. DOI:10.1016/j.eng.2023.01.013
190. Dieber B., Kacianka S., Rass S., et al. (2016). Application-level security for ROS-based applications. *2016 IEEE/RSJ international conference on intelligent robots and systems (IROS)* 4477–4482. DOI:10.1109/iros.2016.7759659
191. DeMarinis N., Tellex S., Kemerlis V. P., et al. (2019). Scanning the internet for ros: A view of security in robotics research. *2019 International conference on robotics and automation (ICRA)* 8514–8521. DOI:10.1109/icra.2019.8794451
192. Chung K., Li X., Tang P., et al. (2019). Smart Malware that Uses Leaked Control Data of Robotic Applications: The Case of Raven- II Surgical Robots. *22nd International Symposium on Research in Attacks, Intrusions and Defenses (RAID 2019)* (*USENIX Association*): 337–351. Available at: https://www.usenix.org/conference/raid2019/presentation/chung
193. Puccetti T., Nardi S., Cinquilli C., et al. (2024). ROSPaCe: Intrusion detection dataset for a ROS2-based cyber-physical system and IoT networks. *Sci. Data* **11**:481. DOI:10.1038/s41597-024-03311-2
194. Mayoral-Vilches V., Pinzger M., Rass S., et al. (2020). Can ROS be used securely in industry? Red teaming ROS-industrial. *arXiv*. DOI: 10.48550/arXiv.2009.08211
195. Qureshi A., Marvi M., Shamsi J. A., et al. (2022). eUF: A framework for detecting over-the-air malicious updates in autonomous vehicles. *J. King Saud Univ. Comput. Inf. Sci.* **34**:5456–5467. DOI:10.1016/j.jksuci.2021.05.005
196. Mocnik R., Fowler D. S. and Maple C. (2023). Vehicular over-the-air software upgrade threat modelling. *Cenex-LCV and cenex-CAM 2023*. Available at: https://wrap.warwick.ac.uk/179188/
197. Mayoral-Vilches V., Ayucar-Carbajo U., Laflamme O., et al. (2026). Cybersecurity AI: Hacking consumer robots in the AI era. *arXiv*. DOI: 10.48550/arXiv.2603.08665
198. Wang X., Pan H., Zhang H., et al. (2024). Robot Collapse: Supply Chain Backdoor Attacks Against VLM-based Robotic Manipulation. *arXiv*. DOI: 10.48550/arXiv.2411.11683
199. Zhou X., Tie G., Zhang G., et al. (2025). BadVLA: Towards backdoor attacks on vision-language-action models via objective-decoupled optimization. *arXiv*. DOI: 10.48550/arXiv.2505.16640
200. Burbano L., Ortiz D., Sun Q., et al. (2025). CHAI: Command Hijacking against

embodied AI. *arXiv*. DOI: 10.48550/arXiv.2510.00181
201. Geng T., Qu Y. and Wong W. E. (2026). A white-box prompt injection attack on embodied AI agents driven by large language models. *J. Syst. Softw.* 235:112782. DOI:10.1016/j.jss.2026.112782
202. Qin X., Luan S., See J., et al. (2026). Harnessing embodied agents: Runtime governance for policy-constrained execution. *arXiv*. DOI: 10.48550/arXiv.2604.07833
203. Onik A. R., Alsmadi R., Baggili I., et al. (2024). So fresh, so clean: Cloud forensic analysis of the Amazon iRobot Roomba vacuum. *Forensic Sci. Int. Digit. Investig.* 48:301686. DOI:10.1016/j.fsidi.2023.301686
204. Ulsmg B., Lin J. C. and Lee M. C. (2024). Investigating the privacy risk of using robot vacuum cleaners in smart environments. *International conference on information and communications security* 312–330. DOI:10.1007/978-981-97-8798-2_16
205. Yang M., Huang C., Huang X., et al. (2025). Privacy-preserved visual simultaneous localization and mapping based on a dual-component approach. *Appl. Sci.* 15:2583. DOI:10.3390/app15052583
206. Ravichandran Z., Robey A., Kumar V., et al. (2026). Safety guardrails for LLM-enabled robots. *IEEE Robot. Autom. Lett.* 11:4649–4656. DOI:10.1109/lra.2026.3667488
207. Huang X., Zhang R., Cheng L., et al. (2026). LLM-guided safety agent for edge robotics with an ISO-compliant perception-compute-control architecture. *arXiv*. DOI: 10.48550/arXiv.2604.20193
208. Obi I., Venkatesh V. L., Wang W., et al. (2026). Pre-execution safety gate & task safety contracts for LLM-controlled robot systems. *arXiv*. DOI: 10.48550/arXiv.2604.05427
209. Liu J., Corbett-Davies J., Ferraiuolo A., et al. (2018). Secure autonomous cyber-physical systems through verifiable information flow control. *Proceedings of the 2018 workshop on cyber-physical systems security and PrivaCy* 48–59. DOI:10.1145/3264888.3264889
210. Astorga A., Hsieh C., Madhusudan P., et al. (2023). Perception contracts for safety of ML-enabled systems. *Proc. ACM Program. Lang.* 7:2196–2223. DOI:10.1145/3622875
211. Zhan S. S., Wang P., Liu Y., et al. (2025). SENTINEL: A Multi-Level Formal Framework for Safety Evaluation of Foundation Model-based Embodied Agents. *arXiv*. DOI: 10.48550/arXiv.2510.12985
212. Yu D., Shi J., Ren J., et al. (2025). Enhancing security in embodied intelligence: Attack detection via constraint functions. *International conference on algorithms and architectures for parallel processing* 96–110. DOI:10.1007/978-981-95-8417-8_8
213. Choi H., Lee W. C., Aafer Y., et al. (2018). Detecting attacks against robotic vehicles: A control invariant approach. *Proceedings of the 2018 ACM SIGSAC conference on computer and communications security* 801–816. DOI:10.1145/3243734.3243752
214. Hobbs K. L., Mote M. L., Abate M. C., et al. (2023). Runtime assurance for safety-critical systems: An introduction to safety filtering approaches for complex control systems. *IEEE Control Syst.* 43:28–65. DOI:10.1109/mcs.2023.3234380
215. Hsu K. C., Hu H. and Fisac J. F. (2024). The safety filter: A unified view of safety-critical control in autonomous systems. *Annu. Rev. Control Robot. Auton. Syst.* 7:47–72. DOI:10.1146/annurev-control-071723-102940
216. Schilliger J., Lew T., Richards S. M., et al. (2021). Control barrier functions for cyber-physical systems and applications to NMPC. *IEEE Robot. Autom. Lett.* 6:8623–8630. DOI:10.1109/lra.2021.3111010
217. Knoedler L., So O., Yin J., et al. (2025). Safety on the fly: Constructing robust safety filters via policy control barrier functions at runtime. *IEEE Robot. Autom. Lett.* 10:10058–10065. DOI:10.1109/lra.2025.3597847
218. Zhang Z., Lei L., Wu L., et al. (2024). SafetyBench: Evaluating the safety of large language models. *Proceedings of the 62nd Annual Meeting of the Association for Computational Linguistics (Volume 1: Long Papers)* 15537–15553. DOI:10.18653/v1/2024.acl-long.830
219. Zhu Z., Wu B., Zhang Z., et al. (2024). EARBench: Towards evaluating physical risk awareness for task planning of foundation model-based embodied AI agents. *arXiv*. DOI: 10.48550/arXiv.2408.04449
220. Sun Q., Chi X., Rui Y., et al. (2026). LABSHIELD: A multimodal benchmark for safety-critical reasoning and planning in scientific laboratories. *arXiv*. DOI: 10.48550/arXiv.2603.11987
221. Lu X., Chen Z., Hu X., et al. (2026). IS-Bench: Evaluating interactive safety of VLM-driven embodied agents in daily household tasks. *Proceedings of the AAAI conference on artificial intelligence* 40:35680–35688. DOI:10.1609/aaai.v40i42.40880
222. Tomilin T., Fang M. and Pechenizkiy M. (2025). HASARD: A benchmark for vision-based safe reinforcement learning in embodied agents. *The thirteenth international conference on learning representations.*
223. Fei S., Wang S., Shi J., et al. (2025). Libero-Plus: In-depth robustness analysis of vision-language-action models. *arXiv*. DOI: 10.48550/arXiv.2510.13626
224. Huang Y., Wang Z., Wan Z., et al. (2025). Annie: Be careful of your robots. *arXiv*. DOI: 10.48550/arXiv.2509.03383
225. Huang Y., Ding L., Tang Z., et al. (2025). A framework for benchmarking and aligning task-planning safety in LLM-based embodied agents. *arXiv*. DOI: 10.48550/arXiv.2504.14650
226. Advanced Technology Exploration Community (2026). ATEC2026: The real-world extreme challenge competition for embodied intelligence. (*Advanced Technology Exploration Community*). Available at: https://www.atecup.com/competitions/ATEC2026
227. DARKNAVY (2026). Embodied AI security technology white paper: Robotics chapter. (*DARKNAVY*). Available at: https://www.darknavy.org/zh/blog/embodied-ai-security-humanoid-robots/
228. Man Y., Muller R., Li M., et al. (2023). That person moves like a car: Misclassification attack detection for autonomous systems using spatiotemporal consistency. *32nd USENIX Security Symposium (USENIX Security 23)* (*USENIX Association*): 6929–6946. Available at: https://www.usenix.org/conference/usenixsecurity23/presentation/man
229. Cao Y., Bhupathiraju S. H., Naghavi P., et al. (2023). You Can't See Me: Physical Removal Attacks on LiDAR-based Autonomous Vehicles Driving Frameworks. *32nd USENIX Security Symposium (USENIX Security 23)* (*USENIX Association*): 2993–3010. Available at: https://www.usenix.org/conference/usenixsecurity23/presentation/cao .
230. Xiao Q., Pan X., Lu Y., et al. (2023). Exorcising "Wraith": Protecting LiDAR-based Object Detector in Automated Driving System from Appearing Attacks. *32nd USENIX Security Symposium (USENIX Security 23)* (*USENIX Association*): 2939–2956. Available at: https://www.usenix.org/conference/usenixsecurity23/presentation/xiao-qifan .
231. Sathaye H., LaMountain G., Closas P., et al. (2022). SemperFi: Anti-spoofing GPS receiver for UAVs. *Network and distributed system security symposium (NDSS 2022)* (*Internet Society*). DOI: 10.14722/ndss.2022.23071
232. Jeong J., Kim D., Jang J. H., et al. (2023). Un-Rocking Drones: Foundations of acoustic injection attacks and recovery thereof. *Network and distributed system security symposium (NDSS 2023)* (*Internet Society*). DOI: 10.14722/ndss.2023.24112
233. Ding A., Murthy P., Garcia L., et al. (2021). Mini-Me, You Complete Me! Data-Driven Drone Security via DNN-based Approximate Computing. *Proceedings of the 24th International Symposium on Research in Attacks, Intrusions and Defenses (RAID)* (*Association for Computing Machinery*): 428–441. DOI: 10.1145/3471621.3471869
234. Sciangula G., Casini D., Biondi A., et al. (2023). Bounding the Data-Delivery Latency of DDS Messages in Real-Time Applications. Papadopoulos A. V. (ed). *35th Euromicro Conference on Real-Time Systems (ECRTS 2023)* (*Schloss Dagstuhl – Leibniz-Zentrum für Informatik*), Leibniz International Proceedings in Informatics (LIPIcs) 262: 9: 1-9: 26. DOI: 10.4230/LIPIcs.ECRTS.2023.9
235. Gu Q., Ju Y., Sun S., et al. (2025). SAFE: Multitask failure detection for vision-language-action models. *Adv. Neural Inf. Process. Syst.* 38. DOI:10.48550/arXiv.2506.09937
236. Zhang B., Zhang Y., Ji J., et al. (2025). SafeVLA: Towards safety alignment of vision-language-action model via constrained learning. Adv. Neural Inf. Process. Syst. 38. Available at: https://neurips.cc/virtual/2025/loc/san-diego/poster/116975
237. Ichnowski J., Chen K., Dharmarajan K., et al. (2023). FogROS2: An adaptive platform for cloud and fog robotics using ROS 2. *IEEE international conference on robotics and automation, ICRA 2023, London, UK, May 29 - June 2, 2023* (*IEEE*): 5493–5500. DOI: 10.1109/ICRA48891.2023.10161307

## FUNDING AND ACKNOWLEDGMENTS

This work was supported by the State Key Lab of Intelligent Transportation System under Project No. 2026-B016; the Zhejiang Provincial Natural Science Foundation of China (No. LR26F020002); the National Natural Science Foundation of China (No. 62672277); the Shandong Provincial Natural Science Foundation (No. ZR2026ZD40); the Key Technology Research and Industrialization Demonstration Projects in Qingdao (No. 26-1-1-gjgg-30-gx); and the Research Project of Quancheng Laboratory, China (No. QCL20250106). The funders had no role in study design, data collection and analysis, decision to publish, or preparation of the manuscript.

## AUTHOR CONTRIBUTIONS

W. Wang, Yue Zhang, S. Zhou, and Z. Fan conceived and designed the study and developed its core concepts and framework. T. Feng conducted the literature review and drafted the ``Introduction'' section, the ``Hardware Safety and Security in Embodied AI'' section, and the ``Embodied Security and Safety Benchmarks'' section. Z. Fan contributed to the literature review and drafted the ``Background'' section. S. Zhou performed the literature review and drafted the ``Software Safety and Security in Embodied AI'' section. Yue Zhang contributed to the literature review and drafted the ``Compound Safety and Security Risks under Software–Hardware Coupling in Embodied AI'' section and the ``Future Directions and Outlook'' section. X. Chi, T. Shen, X. Li, X. Bai, Z. Zhu, S. Han, B. Zhou, Z. Wang, S. Chen, Z. Zheng, C. Bai, J. Pang, Y. Yang, H. Zhao, D. Hu, L. Yi, Y. Huang, Q. Wu, Hanwang Zhang, F. Gao, Shanghang Zhang, H. Li, J. Hao, Yuan Zhang, Y. Xia, J. Shen, C. Lu, Hui Zhang, K. Chen, W. Yu, B. Chen, J. Han, and Y. Guo contributed to the review, revision, and refinement of the manuscript across multiple versions, providing critical feedback and discussions. All authors contributed to the manuscript and approved the final version.

## DECLARATION OF INTERESTS

Shiji Zhou and Wenguan Wang are Editorial Board members of *AI Plus* and were blinded from reviewing or making final decisions on the manuscript. Peer review was handled independently of these members and their research group. The other authors declare no conflicts of interest.

## ETHICAL STATEMENT AND PATIENT CONSENT

Not applicable.

## DATA AND CODE AVAILABILITY

No new data were created or analyzed in this study.

## SUPPLEMENTAL INFORMATION

It can be found online at https://doi.org/10.59717/ipj.aiplus.2026.100013

## LEAD CONTACT WEBSITE

Wenguan Wang: https://sites.google.com/view/wenguanwang